\documentclass[a4paper,fleqn]{cas-dc}

\usepackage[numbers,sort&compress]{natbib}

\usepackage{subcaption}
\usepackage{float}
\usepackage{amsmath}
\usepackage[ruled]{algorithm2e}
\usepackage{soul}
\usepackage{bm}
\usepackage{color, xcolor}
\usepackage{colortbl}
\definecolor{tabred}{HTML}{C83030}
\definecolor{tabblue}{HTML}{3068B0}
\soulregister\cite7
\soulregister\citep7
\soulregister\ref7
 
\usepackage{tabularx}
\usepackage{graphicx}
\usepackage{diagbox}
\usepackage{multirow}
\usepackage{amssymb}
\usepackage{amsthm}
\usepackage{enumitem}

\theoremstyle{plain}

\theoremstyle{definition}

\theoremstyle{remark}

\def\tsc#1{\csdef{#1}{\textsc{\lowercase{#1}}\xspace}}
\tsc{WGM}
\tsc{QE}
\begin{document}
\let\WriteBookmarks\relax
\def\floatpagepagefraction{1}
\def\textpagefraction{.001}

\shorttitle{}
\shortauthors{}

\title [mode = title]{AdaptOVCD: Training-Free Open-Vocabulary Remote Sensing Change Detection via Adaptive Information Fusion}

\author[1,2]{Mingyu Dou}
\author[1,2]{Shi Qiu}\cormark[1]\ead{qiushi@opt.ac.cn}
\author[1,2]{Ming Hu}
\author[3,4]{Yifan Chen}
\author[5,6]{Huping Ye}
\author[5,6]{Xiaohan Liao}
\author[3,7]{Zhe Sun}\cormark[1]\ead{sunzhe@nwpu.edu.cn}

\affiliation[1]{organization={Key Laboratory of Spectral Imaging Technology CAS, Xi'an Institute of Optics and Precision Mechanics, Chinese Academy of Sciences (CAS)},
            city={Xi'an},
            postcode={710119},
            country={China}}
\affiliation[2]{organization={University of Chinese Academy of Sciences},
            city={Beijing},
            postcode={100408},
            country={China}}
\affiliation[3]{organization={Institute of Artificial Intelligence (TeleAI), China Telecom},
            city={Shanghai},
            postcode={200232},
            country={China}}
\affiliation[4]{organization={College of Future Information Technology, Fudan University},
            city={Shanghai},
            postcode={200433},
            country={China}}
\affiliation[5]{organization={State Key Laboratory of Resources and Environment Information System, Institute of Geographic Sciences and Natural Resources Research, CAS},
            city={Beijing},
            postcode={100101},
            country={China}}
\affiliation[6]{organization={Key Laboratory of Low Altitude Geographic Information and Air Route, CAAC},
            city={Beijing},
            postcode={100101},
            country={China}}
\affiliation[7]{organization={School of Artificial Intelligence, Optics and Electronics (iOPEN), Northwestern Polytechnical University},
            city={Xi'an},
            postcode={710072},
            country={China}}
  
\cortext[1]{Corresponding authors.}

\begin{abstract}
Remote sensing change detection plays a pivotal role in domains such as environmental monitoring, urban planning, and disaster assessment. However, existing methods typically rely on predefined categories and large-scale pixel-level annotations, which limit their generalization and applicability in open-world scenarios. To address these limitations, this paper proposes AdaptOVCD, a training-free Open-Vocabulary Change Detection (OVCD) architecture based on dual-dimensional multi-level information fusion. The framework integrates multi-level information fusion across data, feature, and decision levels vertically while incorporating targeted adaptive designs horizontally, achieving deep synergy among heterogeneous pre-trained models to effectively mitigate error propagation. Specifically, (1) at the data level, Adaptive Radiometric Alignment (ARA) fuses radiometric statistics with original texture features and synergizes with SAM-HQ to achieve radiometrically consistent segmentation; (2) at the feature level, Adaptive Change Thresholding (ACT) combines global difference distributions with edge structure priors and leverages DINOv3 to achieve robust change detection; (3) at the decision level, Adaptive Confidence Filtering (ACF) integrates semantic confidence with spatial constraints and collaborates with DGTRS-CLIP to achieve high-confidence semantic identification. Comprehensive evaluations across nine scenarios demonstrate that AdaptOVCD detects arbitrary category changes in a zero-shot manner, significantly outperforming existing training-free methods. Meanwhile, it achieves 84.89\% of the fully-supervised performance upper bound in cross-dataset evaluations and exhibits superior generalization capabilities. The code is available at \url{https://github.com/Dmygithub/AdaptOVCD}.
\end{abstract}

\begin{highlights}
\item Integrates SAM-HQ, DINOv3, and DGTRS-CLIP for training-free open-vocabulary change detection
\item Targeted adaptive designs enable deep synergy among heterogeneous pre-trained models to mitigate error propagation
\item Eliminates reliance on annotations and training, enabling text-driven detection of arbitrary remote sensing change categories
\item Demonstrates exceptional generalization and robustness across nine change detection scenarios
\end{highlights}

\maketitle

\section{Introduction}
\begin{figure*}[t]
  \centering
  \includegraphics[width=\linewidth]{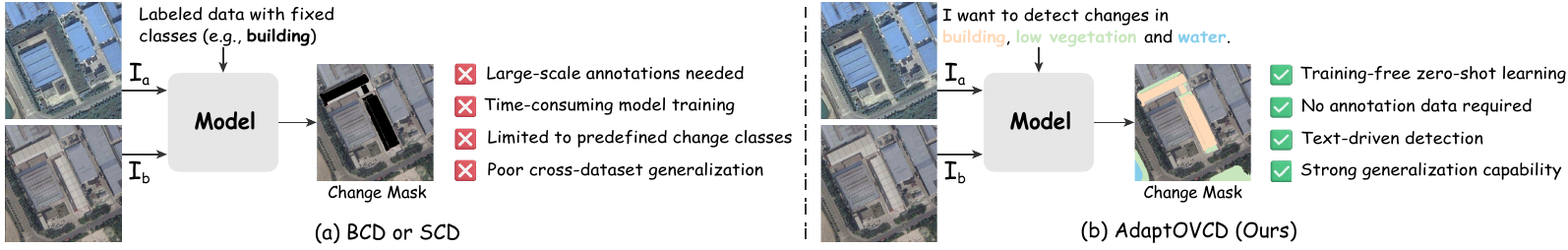}
  \caption{Comparison of change detection paradigms. (a) Traditional supervised methods including BCD and SCD rely on large-scale pixel-level annotations and time-consuming training, restricting their applicability to predefined categories and suffering from limited cross-dataset generalization. (b) The proposed AdaptOVCD framework enables training-free zero-shot detection through natural language prompts, eliminating the need for annotated data while achieving robust generalization across diverse remote sensing scenarios.}
  \label{fig1}
\end{figure*}

Dynamic surface evolution is a central theme in Earth system science. Precise monitoring of spatiotemporal changes is crucial for understanding the environmental impact of human activities and tracking ecosystem evolution. As a key technical approach, Remote Sensing Change Detection (RSCD) identifies surface variations by analyzing bi-temporal images of the same geographic area, playing a pivotal role in urban expansion monitoring, disaster assessment, and land cover mapping~\cite{LV2025103257,dou2025mcamamba,LI2024102240}. The advent of deep learning has established two mainstream paradigms. Binary Change Detection (BCD)~\cite{ZHANG2025103276} aims to output binary masks to localize changes, while Semantic Change Detection (SCD)~\cite{LIU202573} further identifies specific land cover categories. However, as illustrated in Figure~\ref{fig1}(a), these supervised paradigms face inherent limitations, primarily reliance on large-scale pixel-level annotations and extensive training overhead. Furthermore, their restriction to predefined categories limits the detection of unseen classes and hinders cross-dataset generalization.

Recently, breakthroughs in visual foundation models have offered new avenues for addressing these challenges. Through large-scale pre-training, models such as the Segment Anything Model (SAM)~\cite{kirillov2023segany}, Self-Distillation with No Labels (DINO)~\cite{caron2021emerging}, and Contrastive Language-Image Pre-training (CLIP)~\cite{radford2021learning} have demonstrated powerful generalization capabilities in spatial segmentation, semantic representation, and cross-modal understanding, respectively. The complementary strengths of these models enable complex visual understanding without task-specific training. In this context, Open-Vocabulary Change Detection (OVCD) has emerged, aiming to identify changes in arbitrary categories driven by natural language descriptions~\cite{li2025dynamicearth}.

To realize OVCD, researchers have extensively explored the potential of visual foundation models. Existing methodologies primarily follow two technical paradigms. The first category comprises training-based methods~\cite{Zhu2025SemanticCD,ma2025vilacdr1,DONG202453}, which enhance open-vocabulary recognition capabilities by updating model parameters. The second category consists of training-free methods~\cite{li2025dynamicearth,NEURIPS2024_94154162,10642429}, which aim to achieve efficient zero-shot detection through inference-time strategy combinations, keeping pre-trained model parameters frozen. These methods typically integrate multiple pre-trained models to preserve the generalization capability of the foundation models. Currently, training-based approaches are constrained by the scarcity of high-quality annotated data for OVCD and are prone to compromising the inherent generalization capabilities of foundation models. In contrast, training-free methods have emerged as a more viable technical pathway, offering rapid deployment capabilities independent of domain-specific annotations.
 
Nevertheless, existing training-free methods~\cite{NEURIPS2024_94154162,10642429,li2025dynamicearth} still encounter three core challenges. First, task decomposition and collaboration remain a significant hurdle. Training-free methods typically decompose the OVCD task into multiple sub-tasks, employing heterogeneous visual models that require efficient semantic interaction and information fusion. Second, a domain awareness gap persists. General-purpose visual models are predominantly pre-trained on single-temporal tasks in natural scenes. Consequently, they lack dynamic modeling capabilities for bi-temporal images and struggle to adapt to the overhead perspective of remote sensing, resulting in a significant semantic gap when directly transferred. Finally, cascaded error propagation poses a critical issue. Heterogeneous models often adopt static cascading strategies, where data-level radiometric discrepancies and feature-level deviations are progressively amplified, ultimately leading to detection failures at the decision layer. Therefore, achieving efficient task collaboration, bridging the domain perception gap, and establishing effective error suppression mechanisms without training represent key difficulties in realizing high-precision training-free OVCD.
 
To address the challenges mentioned above, this paper proposes AdaptOVCD, a training-free Open-Vocabulary Change Detection (OVCD) framework based on dual-dimen\-sional multi-level information fusion. As shown in Figure~\ref{fig1}(b), AdaptOVCD enables the detection of arbitrary remote sensing category changes driven by text without annotation and training for the OVCD task. To address the task decomposition and collaboration challenge, the framework decomposes OVCD into three sub-tasks, namely instance segmentation, feature comparison, and semantic identification. To bridge the domain awareness gap, the framework utilizes general models to process bi-temporal branches independently, avoiding bi-temporal modeling, introduces the domain-specific foundation model DGTRS-CLIP, and designs mutually exclusive detection prompts to adapt to the remote sensing overhead perspective. To mitigate cascaded error propagation, this architecture constructs multi-level information fusion of data, features, and decisions vertically and conducts targeted adaptive designs horizontally, effectively curbing error propagation and achieving deep synergy among heterogeneous models.

Specifically, at the data level, Adaptive Radiometric Alignment (ARA) is designed to fuse the radiometric statistical features and original texture information of bi-temporal images, integrating Segment Anything Model in High Quality (SAM-HQ)~\cite{sam_hq} to achieve radiometrically consistent segmentation. At the feature level, Adaptive Change Thresholding (ACT) is designed to fuse global difference distributions with edge structure priors, leveraging Self-Distillation with No Labels v3 (DINOv3)~\cite{simeoni2025dinov3} to dynamically build robust change decision boundaries. At the decision level, Adaptive Confidence Filtering (ACF) is designed to fuse semantic confidence with spatial constraints, utilizing Dynamic Ground-Truth Remote Sensing CLIP (DGTRS-CLIP)~\cite{11219074} to output high-confidence change detection results.

The main contributions of this paper are summarized as follows:
\begin{enumerate}
\item We propose a dual-dimensional multi-level information fusion architecture, constructing cascade fusion of data, feature, and decision levels vertically, while conducting targeted adaptive designs horizontally, achieving open-vocabulary change detection without annotated data or training.

\item We implement three adaptive designs, namely ARA, ACT, and ACF, at the data, feature, and decision levels, respectively. These designs synergize with SAM-HQ, DINOv3, and DGTRS-CLIP to foster deep collaboration among heterogeneous pre-trained models, thereby effectively suppressing error accumulation.

\item Comprehensive experiments covering nine change detection scenarios show that AdaptOVCD detects arbitrary category changes in a zero-shot manner, significantly outperforming existing training-free methods, and demonstrates generalization capabilities superior to fully supervised models in cross-dataset evaluations.
\end{enumerate}

\section{Related Work}

\subsection{Traditional Change Detection}

The fundamental objective of traditional change detection lies in accurately identifying changing regions from bi-temporal images. Depending on the output format, traditional methods are categorized into Binary Change Detection (BCD) and Semantic Change Detection (SCD), both of which are constrained by predefined category systems.

\textbf{Binary Change Detection.} BCD aims to determine whether each pixel has changed and output a binary mask. The fundamental paradigm of traditional unsupervised methods involves constructing a difference map and performing change determination via threshold segmentation. CVA~\cite{4039609}, which analyzes the magnitude and direction of multi-band spectral difference vectors in the polar domain, represents a seminal work in this field. IRMAD~\cite{4060945} iteratively extracts change information based on multivariate alteration detection and canonical correlation analysis, effectively mitigating radiometric difference interference. PCA-Kmeans~\cite{5196726} combines Principal Component Analysis with K-means clustering to achieve unsupervised change detection via clustering in a reduced-dimensional feature space. ISFA~\cite{isfa} enhances detection robustness by learning temporally invariant slow features. Deep learning has significantly advanced BCD performance. DSFA~\cite{8824216} fuses slow feature analysis with deep networks to achieve change detection without annotations. DCVA~\cite{8608001} addresses multi-change type detection through Deep Change Vector Analysis, extending the representation capability of traditional CVA. ChangeViT~\cite{ZHU2026112539} pioneered the introduction of a pure Vision Transformer architecture into the change detection field, modeling long-range spatial dependencies through self-attention mechanisms, significantly improving detection accuracy for large-scale change areas. AFCN~\cite{ZHOU2026108234} constructs change features from a frequency domain perspective, combined with high-low frequency separation and edge detection auxiliary tasks, achieving excellent performance on building change detection and damage assessment tasks. However, BCD methods can only determine whether a change has occurred but cannot identify change categories, limiting their practical value.

\textbf{Semantic Change Detection.} To characterize the nature of changes beyond binary masks, SCD further identifies the land cover categories of change regions. HGINet~\cite{LONG2024318} introduces a hierarchical graph interaction network to model the temporal correlation between bi-temporal semantic features and difference features via graph convolutional networks. GSTM-SCD~\cite{LIU202573} presents a graph-enhanced spatiotemporal state space model, capturing spatiotemporal dependencies of multi-temporal remote sensing images through a Mamba encoder to achieve refined semantic change identification. ClearSCD~\cite{tang2024clearscd} comprehensively utilizes semantic and change relationships by decoupling semantic segmentation and change detection tasks. CDLNet~\cite{ZHANG2025103276} proposes a lightweight network for joint content awareness and difference transformation, achieving consistent semantic classification and change localization while maintaining a lightweight architecture through spatiotemporal content awareness fusion modules and multi-type temporal difference transformation modules. SFCGNet~\cite{FENG2025} designs a semantic feedback enhancement mechanism and a cross-scale gated fusion module to bridge the gap between deep semantic and shallow detail features through closed-loop feature interaction. AdaptVFMs-RSCD~\cite{JIANG2025304} achieves semantic change detection by fine-tuning SAM and CLIP, representing an early exploration of applying visual foundation models to this task. However, current SCD methods remain limited by predefined category systems and large-scale pixel-level annotations, unable to detect change types unseen in the training set, resulting in restricted generalization capability.

In summary, whether BCD or SCD, existing methods face the fundamental limitation of closed-set recognition, which has driven research into Open-Vocabulary Change Detection.

\subsection{Visual Foundation Models}

In recent years, visual foundation models, exemplified by visual segmentation models, visual representation models, and visual language models, have acquired robust generalization capabilities through large-scale pre-training, opening new avenues for OVCD research. Functionally, visual foundation models relevant to change detection tasks can be categorized into the following three categories.

\textbf{Visual Segmentation Models.} These are designed to delineate visual scenes into semantically consistent independent regions. Recently, prompt-based universal segmentation models have witnessed breakthrough advancements. SAM~\cite{kirillov2023segany}, pre-trained on the SA-1B dataset containing 1.1 billion masks, achieves robust class-agnostic segmentation capabilities through various prompt forms such as points, boxes, and masks, marking a milestone in the field of general segmentation. SAM-HQ~\cite{sam_hq} addresses the limitation of insufficient boundary segmentation accuracy in SAM for complex boundary scenes by introducing a learnable HQ-Token mechanism to inject high-frequency boundary information, significantly enhancing fine-grained boundary segmentation accuracy. SAM2~\cite{ravi2024sam2} extends segmentation capabilities from static images to video sequences, supporting temporally consistent object tracking and segmentation through a streaming memory mechanism. However, segmentation models are inherently limited to spatial partitioning capabilities and cannot determine the semantic categories of regions or their temporal changes.

\textbf{Visual Representation Models.} These focus on learning generalizable visual features from large-scale unlabeled data. The DINO~\cite{caron2021emerging} series employs a self-\hspace{0pt}distillation paradigm, learning semantically consistent feature representations thro\hspace{0pt}ugh online knowledge distillation of teacher-student networks, representing a quintessential example of this direction. The core advantage of DINO resides in the semantic invariance of its features, i.e., producing highly consistent feature representations for the same semantic object under different lighting and viewing conditions. DINOv2~\cite{oquab2023dinov2} is trained on the larger-scale LVD-142M dataset and combines self-distillation with masked image modeling strategies, significantly improving cross-domain generalization and dense prediction performance. DINOv3~\cite{simeoni2025dinov3} further optimizes the spatial resolution and semantic discriminability of features, achieving superior performance on dense prediction tasks. This semantic invariance makes DINO features particularly suitable for cross-temporal semantic comparison tasks, but they lack precise spatial segmentation capabilities and open-vocabulary classification capabilities.

\textbf{Vision-Language Models.} These facilitate semantic alignment between images and text by jointly learning visual and language representations. CLIP~\cite{radford2021learning} is pre-trained on 400 million image-text pairs via contrastive learning, demonstrating powerful zero-shot image recognition capabilities and establishing a new paradigm for vision-language pre-training. Focusing on the remote sensing field, RemoteCLIP~\cite{liu2024remoteclip} fine-tunes CLIP on large-scale remote sensing image-text data, effectively mitigating the performance degradation of general CLIP under overhead perspectives. VLPRSDet~\cite{LIU2025131712} combines visual language models with YOLO detectors to achieve zero-shot detection of remote sensing objects. DGTRS-CLIP~\cite{11219074}, trained on the DGTRSD dataset containing 1.2 million image-text pairs, adopts a dual-granularity alignment strategy at the image and region levels, demonstrating exceptional performance in zero-shot classification and retrieval tasks in remote sensing scenes. However, existing VLMs primarily focus on single-image understanding tasks and lack modeling capabilities for temporal data.

The aforementioned three categories of visual foundation models exhibit distinct complementary strengths. Segmentation models excel at spatial region partitioning, representation models excel at semantic feature extraction, and VLMs excel at open-vocabulary classification. Consequently, effectively leveraging these models to achieve open-vocabulary change detection has emerged as a central challenge in current research.

\subsection{Open-Vocabulary Change Detection}

Open-vocabulary technology seeks to transcend the categorical constraints of closed-set recognition, enabling models to identify novel categories absent from the training set. Research on open-vocabulary techniques in the remote sensing domain, though emergent, has advanced rapidly. Regarding zero-shot learning, Echo~\cite{CHENG2026103952} achieves zero-shot remote sensing image captioning by generating cross-modal features, and VL-ZSDA-RS~\cite{WANG2026132470} applies visual language models to zero-shot domain adaptation classification of remote sensing images. In semantic segmentation tasks, SegEarth-OV~\cite{li2024segearth} proposes a two-stage framework, first using SAM to generate class-agnostic candidate masks, then using CLIP for semantic annotation, achieving training-free open-vocabulary remote sensing semantic segmentation. In object detection tasks, CastDet~\cite{li2024castdet} proposes a category-aware self-training strategy to enhance open-vocabulary detection capabilities, addressing the weak appearance features and arbitrary orientation characteristics of remote sensing images. However, the aforementioned works are predominantly tailored for single-image understanding tasks. Change detection requires simultaneous processing of bi-temporal images and assessment of cross-temporal semantic changes, posing a challenge for the direct transfer of existing methods.

OVCD aims to detect changes across arbitrary categories driven by text descriptions, representing an emerging research direction that overcomes the fixed category limitations of traditional BCD and SCD. Early exploratory efforts attempted to leverage foundation models in change detection scenarios. U-BDD~\cite{zhang2023buildingdamage} proposed a cascaded framework utilizing Grounding DINO, SAM, and CLIP for building damage detection, but it is restricted to damage assessment of a single building category. UCD-SCM~\cite{10642429} implements unsupervised building change detection based on FastSAM and CLIP, which is fundamentally limited to binary detection for specific categories such as buildings. AnyChange~\cite{NEURIPS2024_94154162} proposes a SAM-based training-free change detection method, achieving class-agnostic change segmentation by exploiting bi-temporal semantic similarity in the SAM encoder's latent space. However, this method only outputs binary masks and cannot identify specific change categories. To overcome these limitations and achieve authentic open-vocabulary detection, researchers have begun to explore two technical paradigms, namely training-based methods and training-free methods.

\textbf{Training-Based Methods.} The core strategy involves introducing domain adaptation training, updating all or part of the model parameters to adapt to the unique distribution of remote sensing scenes. Under the Supervised Fine-tuning paradigm, ChangeCLIP~\cite{DONG202453} utilizes CLIP's multimodal representation learning capability to construct a difference feature compensation module to capture bi-temporal semantic changes. Semantic-CD~\cite{Zhu2025SemanticCD} designs bi-temporal CLIP visual encoders, adopting a decoupled multi-task learning strategy to synergistically optimize BCD and SCD objectives. ViLaCD-R1~\cite{ma2025vilacdr1} further introduces reinforcement learning, training VLMs to complete complex block-level bi-temporal reasoning through supervised fine-tuning. In addition, Unsupervised Adaptation methods attempt to optimize adaptation layers using only unsupervised signals. UniVCD~\cite{zhu2025univcd} freezes the backbones of SAM2 and CLIP but still exploits a feature alignment module to bridge heterogeneous feature spaces. Although such methods can improve accuracy through parameter optimization, the training process risks compromising the generalization attributes of foundation models.

\textbf{Training-Free Methods.} These achieve zero-shot detection by combining completely frozen pre-trained models. DynamicEarth~\cite{li2025dynamicearth} formalizes the OVCD task, proposing the MCI framework based on the segment-then-identify paradigm and the IMC framework based on the identify-then-segment paradigm, and constructs a benchmark to evaluate the performance of various model combinations. Such methods maximally retain the generalization potential of foundation models, avoiding reliance on specific annotated data and training processes.

In summary, while training-based methods can improve accuracy through parameter optimization, they not only face the risk of compromising the generalization characteristics of foundation models but also increase the deployment threshold due to the training process. Considering the scarcity of existing OVCD-annotated data, this paper focuses on the training-free paradigm, exploring how to exploit the potential of pre-trained models in open-vocabulary change detection tasks through efficient model collaboration mechanisms. 

\section{Preliminaries}
\label{sec:preliminaries}

\subsection{Task Definition}

We formalize the OVCD task as a zero-shot inference problem. Given a bi-temporal remote sensing image pair $I_a, I_b \in \mathbb{R}^{H \times W \times 3}$ and a predefined semantic category description $\mathcal{T}$, the objective is to construct a mapping function $\mathcal{M}: (I_a, I_b, \mathcal{T}) \rightarrow M \in \{0,1\}^{H \times W}$ while keeping the pre-trained parameters $\Theta_{pre}$ frozen, ensuring that the output mask $M$ accurately delineates the changing regions matching the text description.

To this end, we decouple OVCD into three sub-tasks. This paradigm eliminates reliance on large-scale annotated data, achieves zero-shot inference by integrating advanced vision foundation models, and specifically mitigates cascading errors. Formally, let $\mathcal{F}_{seg}$, $\mathcal{F}_{cmp}$, and $\mathcal{F}_{cls}$ denote the functions for Instance Segmentation, Feature Comparison, and Semantic Identification, respectively. The complete inference process can be formulated as:
\begin{equation}
M = \mathcal{F}_{cls}\left(\mathcal{F}_{cmp}\left(\mathcal{F}_{seg}(I_a), \mathcal{F}_{seg}(I_b)\right), \mathcal{T}\right)
\end{equation}
where (a) $\mathcal{F}_{seg}: I \rightarrow S$ performs class-agnostic instance segmentation, partitioning the image into a set of spatially independent masks. (b) $\mathcal{F}_{cmp}: (S_a, S_b) \rightarrow C$ identifies change candidate regions through semantic feature comparison, distinguishing authentic changes from pseudo-changes. (c) $\mathcal{F}_{cls}: (C, \mathcal{T}) \rightarrow M$ executes open-vocabulary classification on candidate regions based on text prototypes, generating change masks that align with the target semantics.

\subsection{Vision Foundation Models}

This section details the three pre-trained models incorporated into the AdaptOVCD framework.

\textbf{SAM-HQ}~\cite{sam_hq} is a general instance segmentation model supporting Automatic Mask Generation, capable of performing class-agnostic image discretization and decomposing scenes into spatially independent instance sets. Compared to the original SAM, SAM-HQ significantly enhances segmentation accuracy for complex boundaries by introducing a High-Quality Token mechanism, which is crucial for the precise extraction of fine-grained ground objects in remote sensing images. In this paper, we employ SAM-HQ to implement class-agnostic instance segmentation, generating an instance mask set $S = \{m^i\}_{i=1}^{N}$, where $m \in \{0,1\}^{H \times W}$.

\textbf{DINOv3}~\cite{simeoni2025dinov3} is a visual representation model trained via self-supervised learning, proficient in extracting dense features with robust semantic invariance. This implies it remains resilient to appearance discrepancies of the same ground object caused by non-semantic changes such as lighting and seasonal variations, while exhibiting high sensitivity to authentic changes in ground object categories. In this paper, we leverage DINOv3 for class-agnostic feature extraction, obtaining dense feature maps $\mathbf{F} \in \mathbb{R}^{C \times H \times W}$ from bi-temporal images for subsequent cross-temporal semantic comparison and change determination.

\textbf{DGTRS-CLIP}~\cite{11219074} is a vision-language model pre-trained on the DGTRSD dataset containing 1.2 million remote sensing image-text pairs. It demonstrates superior recognition capabilities for ground objects under remote sensing overhead views, effectively bridging the semantic gap between natural scenes and the remote sensing domain. In this paper, we utilize DGTRS-CLIP to execute open-vocabulary semantic recognition, achieving cross-modal information fusion of vision and language by computing the cosine similarity between the visual features of candidate regions and user-defined text prototypes $\mathcal{T}$.

\begin{figure*}[!ht]
\centering
\includegraphics[width=0.95\linewidth]{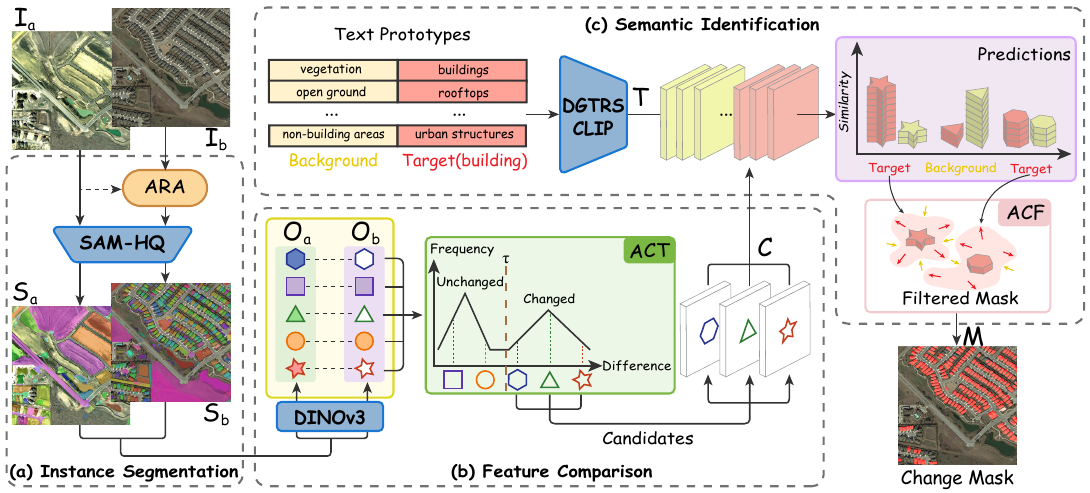}
\caption{Overview of the proposed AdaptOVCD framework for training-free open-vocabulary change detection. The pipeline implements dual-dimensional multi-level information fusion: vertically constructing a data-feature-decision cascade, and horizontally conducting targeted adaptive designs at each level. (a) Instance Segmentation: ARA performs radiometric alignment, followed by SAM-HQ generating class-agnostic masks $S_a$ and $S_b$ from bi-temporal images $I_a$ and $I_b$. (b) Feature Comparison: DINOv3 extracts dense semantic features and computes region-level representations $O_a$ and $O_b$ via mask pooling, while ACT determines adaptive thresholds to identify change candidates $C$. (c) Semantic Identification: DGTRS-CLIP performs open-vocabulary classification using text prototypes $\mathcal{T}$, and ACF applies confidence-based filtering to produce the final change mask $M$.}
\label{fig2}
\end{figure*}

\section{Methodology}
\label{sec:method}

This section systematically details the AdaptOVCD framework. We first present the overall architecture of the framework, followed by a description of the technical details of each stage.

\subsection{AdaptOVCD Framework}

As illustrated in Figure~\ref{fig2}, AdaptOVCD partitions the OVCD task into three stages, achieving deep synergy among heterogeneous models through a dual-dimensional multi-level information fusion architecture. (a) In the instance segmentation stage, ARA fuses radiometric statistical features and raw texture information to ensure radiometric consistency and alignment of bi-temporal images, while SAM-HQ completes class-agnostic instance segmentation. (b) In the feature comparison stage, DINOv3 extracts semantic features of bi-temporal instances, and ACT adaptively determines change decision thresholds by fusing global difference distributions with edge structure priors to filter candidate instances. (c) In the semantic identification stage, DGTRS-CLIP semantically aligns candidate instances with text prototypes, and ACF performs confidence filtering by fusing semantic confidence with spatial connectivity constraints to output high-quality change masks. The processing flow of each stage is detailed below.

\textbf{(a) Instance Segmentation.} This stage ensures radiometrically consistent segmentation through data-level fusion of ARA and SAM-HQ. First, ARA fuses radiometric statistical features and original texture information to adaptively correct the post-temporal image, obtaining the aligned image $I'_b$. Subsequently, SAM-HQ generates sets of spatially independent instance masks $S_a$ and $S_b$ based on the image pair $(I_a, I'_b)$, providing a consistent spatial reference for subsequent analysis. As shown in Figure~\ref{fig2}(a), SAM-HQ performs class-agnostic segmentation on bi-temporal images.

\textbf{(b) Feature Comparison.} This stage facilitates robust change detection through feature-level fusion of DINOv3 and ACT. DINOv3 extracts dense semantic features $\mathbf{F}_a, \mathbf{F}_b \in \mathbb{R}^{C \times H \times W}$ from the aligned image pair $(I_a, I'_b)$. Then, using instance masks $S_a, S_b$, Mask Pooling aggregates the semantic representations of each instance region to obtain region-level features $O_a = \{o_a^i\}_{i=1}^{N_a}$ and $O_b = \{o_b^j\}_{j=1}^{N_b}$. As shown in Figure~\ref{fig2}(b), ACT stratifies instances into unchanged and changed sets based on feature difference distribution, and isolates candidate instances $C$ with significant changes through adaptive thresholds.

\textbf{(c) Semantic Identification.} This stage enables high-confidence semantic recognition through decision-level fusion of DGTRS-CLIP and ACF. As shown in Figure~\ref{fig2}(c), users define mutually exclusive text prototypes $\mathcal{T} = \{T_{target}, \allowbreak T_{background}\}$. DGTRS-CLIP encodes text prototypes into semantic features, calculates the cosine similarity between each candidate mask $m \in C$ and the text prototypes for the target and background classes, respectively, and obtains the probability distribution $p_m$ through softmax normalization. In the Predictions visualization, red and yellow bars represent the similarity of candidate instances to the target class and background class, respectively, with the higher similarity determining the category of the instance. Finally, ACF filters isolated regions with low confidence or abnormal areas by fusing semantic confidence and spatial connectivity constraints, generating precise change masks $M$.

This framework effectively mitigates cascaded error propagation through vertical data-level, feature-level, and decision-level cascade fusion, as well as horizontal targeted adaptive design, achieving robust open-vocabulary change detection in the absence of annotated data. Algorithm~\ref{alg:adaptovcd} presents the complete inference process of the AdaptOVCD framework. The technical details of each stage are elaborated in the following sections.

\begin{algorithm}[!t]
\small{
\caption{AdaptOVCD: Training-Free OVCD}
\label{alg:adaptovcd}
\LinesNumbered
\KwIn{Bi-temporal images $I_a, I_b \in \mathbb{R}^{H \times W \times 3}$; Text prototypes $\mathcal{T} = \{T_{target}, \allowbreak T_{background}\}$; Thresholds $\tau_{max}$}
\KwOut{Binary change mask $M \in \{0,1\}^{H \times W}$}

\textbf{(a) Instance Segmentation} \\
$\tilde{I}_b \gets \text{RadiometricTransfer}(I_b, I_a)$ \tcp*{align $I_b$ to $I_a$}
$\Delta_{max} \gets \max|\tilde{I}_b - I_b| / 255$ \\
\eIf{$\Delta_{max} > \tau_{max}$}{
    $I'_b \gets (\tau_{max}/\Delta_{max}) \cdot \tilde{I}_b + (1 - \tau_{max}/\Delta_{max}) \cdot I_b$
}{
    $I'_b \gets \tilde{I}_b$
}
$S_a \gets \text{SAM-HQ}(I_a, \text{mode}=\text{auto})$ \tcp*{$N_a$ masks}
$S_b \gets \text{SAM-HQ}(I'_b, \text{mode}=\text{auto})$ \tcp*{$N_b$ masks}

\textbf{(b) Feature Comparison} \\
$\mathbf{F}_a \gets \text{Interpolate}(\text{DINOv3}(I_a), (H,W))$ \\
$\mathbf{F}_b \gets \text{Interpolate}(\text{DINOv3}(I'_b), (H,W))$ \\
$S_{all} \gets S_a \cup S_b$ \tcp*{combine instance masks}
$D \gets \text{Normalize}(\|\mathbf{F}_a - \mathbf{F}_b\|_2)$ \tcp*{pixel-wise difference}
$\tau_{global} \gets \text{Otsu}(D)$; $E \gets \text{EdgeDetect}(D)$; $\tau_{edge} \gets \text{Otsu}(D[E])$ \\
$\tau_{final} \gets w_g \cdot \tau_{global} + w_e \cdot \tau_{edge}$ \tcp*{$w_g + w_e = 1$}
$\theta_{threshold} \gets \text{MapToAngle}(\tau_{final})$ \tcp*{linear mapping to angle range}
$C \gets \emptyset$ \\
\For{each mask $m_k \in S_{all}$}{
    $O_a^k \gets \frac{1}{|m_k|}\sum_{(x,y) \in m_k} \mathbf{F}_a[:,x,y]$ \\
    $O_b^k \gets \frac{1}{|m_k|}\sum_{(x,y) \in m_k} \mathbf{F}_b[:,x,y]$ \\
    $s_k \gets O_a^k \cdot O_b^k / (\|O_a^k\| \|O_b^k\|)$ \tcp*{cosine similarity}
    \If{$s_k < \cos(180^\circ - \theta_{threshold})$}{
        $C \gets C \cup \{m_k\}$ \tcp*{add to candidate set}
    }
}

\textbf{(c) Semantic Identification} \\
$M \gets \mathbf{0}^{H \times W}$; $\mathcal{P} \gets \emptyset$ \\
\For{each candidate mask $m \in C$}{
    $p_m \gets \text{Classify}(m, I_b, \mathcal{T})$ \tcp*{get target prob}
    $\mathcal{P} \gets \mathcal{P} \cup \{p_m\}$
}
$\tau_{conf} \gets \text{Percentile}(\mathcal{P}, p) / \lambda$ \tcp*{adaptive threshold}
\For{each candidate mask $m \in C$}{
    $p_m \in \mathcal{P}$ \\
    \If{$p_m > \tau_{conf}$}{
        $M \gets M \cup m$ \tcp*{add to change mask}
    }
}
$M \gets \text{ConnectedFilter}(M)$ \tcp*{spatial-semantic consistency}
\Return{$M$}
}
\end{algorithm}
  
\subsection{Instance Segmentation with Data-Level Fusion Using ARA}

As shown in Figure~\ref{fig2}(a), the data flow in this stage proceeds as $(I_a, I_b) \xrightarrow{\text{ARA}} (I_a, I'_b) \xrightarrow{\text{SAM-HQ}} (S_a, S_b)$. The fundamental objective of this stage is to mitigate non-semantic disparities in bi-temporal images through data-level radiometric correction, providing a consistent radiometric baseline for subsequent segmentation.

\textbf{Adaptive Radiometric Alignment.} ARA addresses the challenge of radiometric discrepancies in bi-temporal remote sensing images. Arising from variations in solar elevation angle, atmospheric conditions, and sensor responses, bi-temporal images often exhibit significant radiometric differences, resulting in inconsistent segmentation outcomes by SAM-HQ across different phases, which in turn propagates data-level bias to subsequent stages and induces cascaded error propagation. Conventional image enhancement techniques, while improving visual aesthetics, tend to alter local texture distributions, compromising the gradient and edge features that SAM relies on. This highlights a critical constraint of the pre-processing strategy within the training-free framework. Specifically, radiometric correction must eliminate non-semantic differences while preserving the visual features utilized by downstream models intact. Guided by this analysis, the design of ARA adheres to the SAM-Safe radiometric correction principle, i.e., maintaining local gradient structures and edge features unchanged while eliminating radiometric differences, to ensure SAM segmentation performance remains unaffected.

Specifically, we formulate a channel-level radiometric transfer function $T_c(v) = (CDF_a^c)^{-1}(CDF_b^c(v))$ to align the intensity distribution of the post-temporal image $I_b$ with that of the pre-temporal image $I_a$. This monotonic mapping maintains the relative grayscale order among pixels, thereby preserving local gradient structures. To prevent information loss induced by over-correction, we further introduce an adaptive mixing strategy modulated by the measured maximum radiometric difference $\Delta_{max}$:
\begin{equation}
\begin{cases}
\alpha = \min(1, \tau_{max} / \Delta_{max}), \\
I'_b = \alpha \tilde{I}_b + (1-\alpha) I_b.
\end{cases}
\end{equation}
where $\tilde{I}_b$ is the result of radiometric transfer, $I'_b$ is the final aligned image, and $\tau_{max}$ is a preset safety threshold. This adaptive mixing strategy synergistically integrates radiometric statistical features with raw texture information, preserving the intrinsic semantic features of the scene while mitigating radiometric differences.

After ARA correction, we employ SAM-HQ to perform class-agnostic instance segmentation on bi-temporal images. SAM-HQ operates in Automatic Mask Generation mode, generating spatially independent instance mask sets $S_a$ and $S_b$ for the image pair $(I_a, I'_b)$ respectively, providing a consistent spatial reference for subsequent feature comparison.

\subsection{Feature Comparison via Feature-Level Fusion Using ACT}
\label{sec:detect}

As shown in Figure~\ref{fig2}(b), this stage aims to identify semantic change candidate regions from instance masks, with the data flow: $(S_a, S_b) \xrightarrow{\text{DINOv3}} (O_a, O_b) \xrightarrow{\text{ACT}} C$. DINOv3 extracts dense semantic features $\mathbf{F}_a, \mathbf{F}_b \in \mathbb{R}^{C \times H \times W}$ from the aligned image pair. We merge the bi-temporal instance masks into $S_{all} = S_a \cup S_b$. For each mask $m_k$, we compute its region features $O_a^k$ and $O_b^k$ at both phases via Mask Pooling, and quantify the magnitude of semantic change using cosine similarity $s_k$:
\begin{equation}
\begin{aligned}
O_t^k &= \frac{1}{|m_k|}\sum_{(x,y) \in m_k} \mathbf{F}_t[:,x,y], \quad t \in \{a, b\} \\
s_{k} &= \frac{O_a^k \cdot O_b^k}{\|O_a^k\| \|O_b^k\|}
\end{aligned}
\end{equation}
This in-situ comparison strategy captures the semantic evolution of the target region in the temporal dimension. Whether it is content replacement or object disappearance, it will manifest as a significant decrease in similarity in the feature space.

\textbf{Adaptive Change Thresholding.} The central challenge of ACT resides in determining the change decision threshold for similarity $s_k$. Since the training-free architecture lacks ground truth labels for learning optimal decision boundaries, it must rely on the distribution characteristics of the data itself for unsupervised change determination. As illustrated in the Frequency curve in Figure~\ref{fig2}(b), the feature difference distribution in change detection scenes typically exhibits bimodal characteristics, corresponding to changed and unchanged regions, respectively. This bimodal characteristic renders the threshold selection strategy based on maximizing inter-class variance inherently suitable for change detection tasks. Accordingly, ACT models the change threshold selection problem as locating the optimal decision boundary differentiating the bimodal distribution and autonomously determines this optimal splitting point.

The technical implementation of ACT includes the following key steps:

\textit{(a) Feature Difference Map Construction.} For bi-temporal DINOv3 features $\mathbf{F}_a, \mathbf{F}_b \in \mathbb{R}^{C \times H \times W}$, we compute the pixel-wise L2 norm difference and perform min-max normalization:
\begin{equation}
D(i,j) = \frac{\|\mathbf{F}_a[:,i,j] - \mathbf{F}_b[:,i,j]\|_2 - D_{min}}{D_{max} - D_{min}} \in [0,1]
\end{equation}
obtaining the feature difference map $D \in \mathbb{R}^{H \times W}$, where larger values indicate more significant semantic changes at the corresponding position.

\textit{(b) Global Threshold.} We apply the maximum inter-class variance strategy on the entire difference map $D$ to automatically find the optimal splitting point $\tau_{global}$.
\begin{equation}
\tau_{global} = \arg\max_{\tau} \sigma^2_B(\tau) = \arg\max_{\tau} w_0(\tau) w_1(\tau) (\mu_0(\tau) - \mu_1(\tau))^2
\end{equation}
where $w_0, w_1$ represent the proportions of pixels on both sides of the threshold, and $\mu_0, \mu_1$ are the corresponding means.

\textit{(c) Edge Region Local Threshold.} Considering that the semantic boundaries of target edge regions are clearer, their feature changes are usually more pronounced and stable than those in internal regions. We apply edge detection on the difference map $D$ to extract the edge mask $E$, and perform moderate morphological dilation to obtain the edge neighborhood. We compute the local threshold $\tau_{edge}$ only on the difference values in the edge region. When the number of edge pixels is insufficient $N_{min}$, $\tau_{edge}$ is set to invalid.

\textit{(d) Threshold Fusion.} The final threshold is obtained by weighted combination:
\begin{equation}
\tau_{final} = \begin{cases}
w_g \tau_{global} + w_e \tau_{edge}, & \text{if } \tau_{edge} \text{ is valid}, \\
\tau_{global}, & \text{otherwise}.
\end{cases}
\end{equation}
where $w_g$ and $w_e$ are adjustment weights, aiming to balance the integrity of the global distribution with the local sensitivity of the edge region. In experiments, these two weights are flexibly adjusted according to the characteristics of the specific scene to achieve the optimal detection performance under different ground object distributions.

To address the discrepancy in metric spaces between the cosine similarity with range $[-1, 1]$ and the difference map statistics with range $[0, 1]$, we linearly map $\tau_{final}$ to the angular space to construct a geometrically consistent decision boundary:
\begin{equation}
\theta_{threshold} = \theta_{min} + \tau_{final}(\theta_{max} - \theta_{min})
\end{equation}
where $\theta_{min}, \theta_{max}$ denote the angle mapping interval. Since the optimal decision boundaries for different ground object categories differ, this interval parameter needs to be adaptively configured according to the specific scene. Under this mapping, a larger $\tau_{final}$ implies a significant scene difference, which will be mapped to a larger $\theta_{threshold}$. Since cosine similarity is negatively correlated with feature difference, the final change decision condition is expressed as
\begin{equation}
s_k < \cos(180^\circ - \theta_{threshold})
\end{equation}
This condition is equivalent to $s_k < -\cos(\theta_{threshold})$, which requires the cosine similarity between features to be lower than the threshold $-\cos(\theta_{threshold})$ to be judged as a change. This mapping mechanism effectively translates the difference distribution characteristics derived from statistics into the similarity decision boundary in geometric space.

\subsection{Semantic Identification via Decision-Level Fusion Using ACF}

As shown in Figure~\ref{fig2}(c), this stage is designed to semantically classify candidate regions and output reliable change masks, with data flow: $C \xrightarrow{\text{DGTRS-CLIP}} \mathcal{P} \xrightarrow{\text{ACF}} M$. We utilize the remote sensing pre-trained DGTRS-CLIP for open-vocabulary classification. Candidate mask $m \in C$ is first cropped to the corresponding region, and after feature extraction by the DGTRS-CLIP visual encoder, cosine similarity is calculated with text prototypes, and semantic probability $p_m$ is obtained through softmax normalization.

Works such as SegEarth-OV~\cite{li2024segearth} that require semantic features and text alignment often adopt mutually exclusive category schemes to enhance discrimination. We design mutually exclusive text prototypes combining the semantic space characteristics of DGTRS-CLIP and the change detection task. Empirical observations indicate that even for models pre-trained on long text descriptions like DGTRS-CLIP, aligning semantics using vocabulary clusters is often superior to long sentences. Therefore, when constructing text prototypes $\mathcal{T} = \{T_{target}, \allowbreak T_{background}\}$, the target prototype employs category-related core vocabulary combinations, while the background prototype explicitly precludes the target category to form strong semantic opposition. For different change detection scenes, we design corresponding mutually exclusive detection texts, details in Table~\ref{tab:prompts}.

\textbf{Adaptive Confidence Filtering.} ACF aims to synergistically address the confidence calibration bias and spatial inconsistency problems. Addressing the intrinsic spatial misalignment of SAM bi-temporal independent segmentation, exacerbated by the potential calibration bias risk of VLM raw predictions, ACF builds an instance-level decision mechanism fusing semantic confidence distribution and spatial connectivity domain geometric constraints to comprehensively ensure output integrity.

Specifically, ACF first adopts an adaptive threshold strategy to deal with the confidence calibration problem. We extract the confidence set $\mathcal{P}_{pos}$ of all instances judged as the target category, take its $p$-th percentile as the baseline, and dynamically calculate the threshold combined with the filtering intensity factor $\lambda$:
\begin{equation}
\tau_{conf} = \text{clip}\left(\text{percentile}(\mathcal{P}_{pos}, p) / \lambda, \, \tau_{min}, \, \tau_{max}\right)
\end{equation}
This strategy can adaptively adjust the filtering boundary according to the confidence distribution characteristics of the current scene. After filtering high-confidence instances using $\tau_{conf}$ and merging to generate a preliminary mask, ACF performs 8-connected component analysis on it to handle the spatial inconsistency of bi-temporal masks, obtaining independent region set $\{R_k\}_{k=1}^{K}$. For each connected domain $R_k$, calculate its mean confidence $\mu_k$ and coefficient of variation $CV_k = \sigma_k / \mu_k$. A region is judged reliable only if it simultaneously satisfies:
\begin{equation}
\lvert R_k \rvert \geq A_{min} \quad \land \quad \mu_k \geq \tau_{min} \quad \land \quad CV_k < \gamma
\end{equation}
where $A_{min}$ is the minimum area threshold used to filter out fragmented artifacts, and $\gamma$ is the upper limit of the coefficient of variation to ensure the consistency of confidence distribution within the region. This judgment is based on the instance level rather than the pixel level. If a region satisfies the above conditions as a whole, all pixels in the region are retained to ensure the spatial integrity of the target. Otherwise, isolated regions with low confidence or unstable confidence are regarded as pseudo-detections caused by misregistration due to parallax and filtered out, finally yielding a high-quality change mask $M$.

\section{Experimental Results and Analysis}

This section first outlines the experimental settings, subsequently benchmarks AdaptOVCD against unsupervised and training-free methods, proceeds to validate open-\hspace{0pt}vocabulary semantic change detection capabilities, analyzes the contributions of each module through ablation studies, and finally verifies the model's generalization ability through cross-dataset evaluation.

\subsection{Datasets}
\textbf{LEVIR-CD Dataset.} The LEVIR-CD~\cite{chen2020levir} dataset comprises 637 pairs of bi-temporal high-resolution remote sensing images, with image dimensions of 1024$\times$1024 pixels and a spatial resolution of 0.5 meters. The dataset documents building alterations in multiple urban areas in China from 2002 to 2018, mainly including new construction, demolition, and expansion of buildings. It is partitioned into 445 pairs for training, 64 pairs for validation, and 128 pairs for testing, providing a large-scale benchmark for building change detection tasks.

\textbf{WHU-CD Dataset.} The WHU-CD~\cite{ji2018whucd} dataset, released by Wuhan University, consists of a pair of high-resolution aerial images covering the Christchurch area in New Zealand, with dimensions of 32507$\times$15354 pixels and a spatial resolution of 0.2 meters. This dataset is cropped into 7434 training pairs and 744 testing pairs of 256$\times$256 pixels, offering authentic application data for earthquake-induced building damage detection.

\textbf{DSIFN Dataset.} The DSIFN~\cite{zhang2020dsifn} dataset includes 3940 pairs of bi-temporal high-resolution satellite images, with dimensions of 512$\times$512 pixels and a spatial resolution of approximately 2 meters. The dataset spans multiple urban areas in China from 2000 to 2019. It is divided into 3600 pairs for training, 340 pairs for validation, and 48 pairs for testing, focuses on building change detection, and is suitable for evaluating model generalization in different urban scenes.

\textbf{SECOND Dataset.} The SECOND~\cite{yang2021second} dataset is tailored for multi-class semantic change detection tasks, containing 4662 pairs of bi-temporal remote sensing images with dimensions of 512$\times$512 pixels and spatial resolutions ranging from 0.5 to 3 meters. The dataset is divided into 2968 pairs for training and 1694 pairs for testing, delineating change regions for six land-cover categories. SECOND provides multi-class semantic annotations, supporting fine-grained change type identification and evaluation.

\begin{table}[t]
\centering
\caption{Pre-trained model configurations used in comparative experiments and architecture ablation studies. Model parameters and storage sizes are reported for fair comparison. SAM-series models uniformly adopt ViT-H backbone, while DINO-series models adopt ViT-L backbone to ensure comparable computational costs.}
\label{tab:implementation}
\renewcommand{\arraystretch}{1.15}
\resizebox{\columnwidth}{!}{
\begin{tabular}{llccc}
\toprule
\textbf{Method} & \textbf{Model} & \textbf{Backbone} & \textbf{Params} & \textbf{Size} \\
\midrule
AnyChange & SAM & ViT-H & 637M & 2.4 GB \\
\midrule
DCVA & Custom ResNet & ResNet-9 & 10M & 40 MB \\
\midrule
\multirow{2}{*}{UCD-SCM} & FastSAM-X & YOLOv8x-seg & 68M & 138 MB \\
 & CLIP & ViT-B/32 & 151M & 338 MB \\
\midrule
\multirow{2}{*}{DynamicEarth (IMC)} & APE & Swin-L & 284M & 1.1 GB \\
 & DINOv2 & ViT-L/14 & 300M & 1.1 GB \\
\midrule
\multirow{3}{*}{DynamicEarth (MCI)} & SAM & ViT-H & 637M & 2.4 GB \\
 & DINOv2 & ViT-L/14 & 300M & 1.1 GB \\
 & SegEarth-OV & ViT-B/16 & 150M & 580 MB \\
\midrule
\multirow{3}{*}{AdaptOVCD (Ours)} & SAM-HQ & ViT-H & 637M & 2.4 GB \\
 & DINOv3 & ViT-L/16 & 300M & 1.1 GB \\
 & DGTRS-CLIP & ViT-L/14 & 428M & 1.7 GB \\
\midrule
\multirow{3}{*}{AdaptOVCD (Variants)} & SAM & ViT-H & 637M & 2.4 GB \\
 & DINOv2 & ViT-L/14 & 300M & 1.1 GB \\
 & DINO & ViT-L/16 & 300M & 1.1 GB \\
\bottomrule
\multicolumn{5}{l}{\textbf{Hardware:} NVIDIA RTX 3090, 24GB, PyTorch.} \\
\multicolumn{5}{l}{For fair comparison, SAM-series use ViT-H and DINO-series use ViT-L uniformly.} \\
\end{tabular}}
\end{table}

\subsection{Experimental Setup}
\textbf{Implementation.} The AdaptOVCD framework is built upon PyTorch, and all experiments are conducted on an NVIDIA RTX 3090 GPU with 24GB VRAM. Table~\ref{tab:implementation} outlines the pre-trained model weight configurations used in subsequent comparative experiments and architecture ablation studies. To facilitate equitable comparison, models of similar scale are used where possible, with the SAM series uniformly employing the ViT-H backbone and the DINO series employing the ViT-L backbone. Notably, AdaptOVCD supports arbitrary resolution inputs. SAM-HQ and DINOv3 natively support arbitrary sizes, while DGTRS-CLIP performs inference on cropped local regions, ensuring the framework can adapt to remote sensing images of any scale.

\textbf{Hyperparameter Configuration.} Considering the differences in spatial resolution and imaging conditions across scenarios, hyperparameters for each module are dynamically tailored to fully leverage the potential of the multi-level fusion mechanism. These hyperparameters include SAM segmentation density and confidence thresholds, ACT module global and edge weights, angle mapping intervals, and ACF module region area and filtering intensity. 

\begin{table*}[t]
\centering
\caption{Text prompt configurations for each dataset. Target prototypes describe the change category of interest using core semantic keywords, while background prototypes explicitly exclude target categories to establish discriminative decision boundaries.}
\label{tab:prompts}
\renewcommand{\arraystretch}{1.3}
\resizebox{\textwidth}{!}{
\begin{tabular}{lll}
\toprule
\textbf{Dataset} & \textbf{Background} & \textbf{Target} \\
\midrule
\textbf{LEVIR-CD} & \makecell[l]{open ground, vegetation,\\and non-building areas} & \makecell[l]{buildings, rooftops,\\and urban structures} \\
\hline
\textbf{WHU-CD} & \makecell[l]{background, open ground, trees, vegetation,\\green spaces, roads, non-building areas} & \makecell[l]{large urban buildings, large rooftops,\\urban houses, urban structures} \\
\hline
\textbf{DSIFN} & \makecell[l]{open areas\\and vegetation} & \makecell[l]{roofs\\and constructions} \\
\midrule
\textbf{Building on SECOND} & \makecell[l]{open ground, vegetation,\\and non-building areas} & \makecell[l]{urban structures,\\rooftops, buildings} \\
\hline
\textbf{Water on SECOND} & \makecell[l]{background, vegetation, land,\\roads, buildings, soil} & \makecell[l]{aquatic areas, waterways, wetlands,\\rivers, lakes, ponds, streams} \\
\hline
\textbf{Tree on SECOND} & \makecell[l]{non-natural areas,\\background} & \makecell[l]{trees, forest,\\woodland, greenery} \\
\hline
\textbf{Low Vegetation on SECOND} & \makecell[l]{bare ground, soil, sand,\\buildings, roads, water} & \makecell[l]{grass, lawn, low vegetation,\\shrubs, ground cover, small plants} \\
\hline
\textbf{Non-veg. on SECOND} & \makecell[l]{buildings, trees, water, vegetation,\\grass, forest, vegetated areas} & \makecell[l]{sand, bare ground, exposed soil, dirt,\\paved surfaces, concrete, asphalt, rocky ground} \\
\hline
\textbf{Playground on SECOND} & \makecell[l]{non-athletic areas,\\background} & \makecell[l]{playgrounds, sports fields,\\game courts, running tracks} \\
\bottomrule
\end{tabular}}
\end{table*}

\textbf{Text Prompts.} Crucial to text-driven change detection is the construction of high-quality semantic descriptions. Although DGTRS-CLIP supports long text input, lengthy descriptions tend to divert the model's focus and dilute the efficacy of feature alignment with core semantics. Therefore, this paper adopts a keyword ensemble strategy, covering diverse visual forms of target objects through concise synonym combinations. As shown in Table~\ref{tab:prompts}, we construct mutually exclusive text prototypes $\mathcal{T} = \{T_{target}, \allowbreak T_{background}\}$. The Target uses category-related core entity vocabulary, while the Background explicitly excludes target categories to form strong discriminative semantic opposition, thereby establishing clear decision boundaries in the DGTRS-CLIP semantic space.

Central to text prototype design is clearly defining the Target and Background with the most concise descriptions. For example, the DSIFN test set is small with a single change category, so a few core words for Target and Background are sufficient. In contrast, WHU-CD covers more diverse building scenes, requiring richer vocabulary combinations. Comparing LEVIR-CD and WHU-CD, the former uses general descriptions for dense small-scale residential buildings at 0.5m resolution, while the latter emphasizes "large urban buildings" and "large rooftops" for large-scale urban buildings at 0.2m resolution to match their significant scale features in overhead views. In building detection, Target consistently uses entity words like "buildings" and "rooftops", while Background uses exclusionary descriptions like "non-building areas" or basic land cover categories like "vegetation" and "open ground" to ensure strong semantic mutual exclusivity.

\begin{table*}[t]
    \centering
    \caption{Quantitative comparison of AdaptOVCD with traditional unsupervised and training-free methods on four building change detection datasets. All metrics are computed on the changed class (denoted by superscript $^C$). Precision (\colorbox{blue!8}{Prec.}), Recall (\colorbox{blue!8}{Rec.}) and \colorbox{blue!8}{IoU$^C$} are standard metrics. \colorbox{orange!10}{F$_1^C$} serves as the primary evaluation metric. \textcolor{tabred}{\textbf{Best}} and \textcolor{tabblue}{\textbf{second best}} results among unsupervised methods are highlighted. The \colorbox{gray!10}{gray row} shows fully-supervised upper bounds for reference only.}
    \renewcommand{\arraystretch}{1.1}
    \resizebox{\textwidth}{!}{
    \begin{tabular}{l|cccc|cccc|cccc|cccc}
    \toprule
    & \multicolumn{4}{c|}{\textbf{LEVIR-CD}} & \multicolumn{4}{c|}{\textbf{WHU-CD}} & \multicolumn{4}{c|}{\textbf{DSIFN}} & \multicolumn{4}{c}{\textbf{Building on SECOND}} \\
    \textbf{Method} & \cellcolor{blue!8}Prec. & \cellcolor{blue!8}Rec. & \cellcolor{blue!8}IoU$^C$ & \cellcolor{orange!10}F$_1^C$ & \cellcolor{blue!8}Prec. & \cellcolor{blue!8}Rec. & \cellcolor{blue!8}IoU$^C$ & \cellcolor{orange!10}F$_1^C$ & \cellcolor{blue!8}Prec. & \cellcolor{blue!8}Rec. & \cellcolor{blue!8}IoU$^C$ & \cellcolor{orange!10}F$_1^C$ & \cellcolor{blue!8}Prec. & \cellcolor{blue!8}Rec. & \cellcolor{blue!8}IoU$^C$ & \cellcolor{orange!10}F$_1^C$ \\
    \midrule \noalign{\vskip -0.35em}
    \multicolumn{17}{l}{\scriptsize\textit{Traditional unsupervised methods}} \\
    CVA~\cite{4039609} & 5.43 & 93.93 & 5.41 & 10.26 & 3.73 & 92.20 & 3.71 & 7.16 & 18.02 & 92.16 & 17.75 & 30.14 & 11.65 & 97.08 & 11.61 & 20.81 \\
    IRMAD~\cite{4060945} & 11.06 & 14.62 & 6.72 & 12.59 & 4.19 & 13.64 & 3.31 & 6.41 & 28.63 & 7.05 & 5.99 & 11.31 & 23.26 & 22.60 & 12.95 & 22.93 \\
    PCA-Kmeans~\cite{5196726} & 5.92 & 36.13 & 5.36 & 10.18 & 7.24 & 44.72 & 6.64 & 12.46 & 27.28 & 43.16 & 20.07 & 33.43 & 17.74 & 44.32 & 14.51 & 25.34 \\
    ISFA~\cite{isfa} & 5.37 & \textcolor{tabred}{\textbf{96.27}} & 5.36 & 10.17 & 3.83 & 92.99 & 3.82 & 7.36 & 17.51 & \textcolor{tabred}{\textbf{96.81}} & 17.41 & 29.66 & 11.46 & \textcolor{tabred}{\textbf{98.49}} & 11.44 & 20.52 \\
    DSFA~\cite{8824216} & 8.90 & 41.75 & 7.92 & 14.67 & 6.67 & 33.99 & 5.91 & 11.15 & 26.98 & 35.91 & 18.21 & 30.81 & 20.77 & 43.48 & 16.35 & 28.11 \\
    DCVA~\cite{8608001} & 18.72 & 35.46 & 13.28 & 24.50 & 21.35 & 42.18 & 15.62 & 28.36 & 32.47 & 28.63 & 17.85 & 30.42 & 25.18 & 38.74 & 18.23 & 30.52 \\
    \midrule \noalign{\vskip -0.35em}
    \multicolumn{17}{l}{\scriptsize\textit{Training-free methods}} \\
    AnyChange~\cite{NEURIPS2024_94154162} & 11.96 & 18.50 & 7.85 & 14.56 & 11.13 & 17.80 & 7.30 & 13.59 & 16.49 & 31.50 & 12.14 & 21.64 & 11.54 & 17.68 & 7.51 & 13.96 \\
    UCD-SCM~\cite{10642429} & 23.99 & 48.73 & 19.15 & 32.15 & 23.40 & 73.47 & 21.58 & 35.50 & 55.06 & 16.30 & 14.38 & 25.15 & 31.99 & 40.61 & 21.79 & 35.79 \\
    DynamicEarth (IMC)~\cite{li2025dynamicearth} & \textcolor{tabblue}{\textbf{62.15}} & 70.18 & \textcolor{tabblue}{\textbf{49.68}} & \textcolor{tabblue}{\textbf{65.78}} & \textcolor{tabblue}{\textbf{67.32}} & 72.24 & \textcolor{tabblue}{\textbf{53.76}} & \textcolor{tabblue}{\textbf{69.68}} & 48.23 & 37.54 & 26.13 & 42.18 & 52.50 & 45.10 & 32.03 & 48.51 \\
    DynamicEarth (MCI)~\cite{li2025dynamicearth} & 47.71 & 59.69 & 36.08 & 53.03 & 42.27 & \textcolor{tabred}{\textbf{83.71}} & 39.06 & 56.17 & \textcolor{tabblue}{\textbf{58.52}} & 51.76 & \textcolor{tabblue}{\textbf{37.89}} & \textcolor{tabblue}{\textbf{54.98}} & \textcolor{tabblue}{\textbf{58.80}} & 50.50 & \textcolor{tabblue}{\textbf{37.76}} & \textcolor{tabblue}{\textbf{54.28}} \\
    AdaptOVCD (Ours) & \textcolor{tabred}{\textbf{62.83}} & \textcolor{tabblue}{\textbf{74.10}} & \textcolor{tabred}{\textbf{51.52}} & \textcolor{tabred}{\textbf{68.00}} & \textcolor{tabred}{\textbf{80.60}} & \textcolor{tabblue}{\textbf{72.85}} & \textcolor{tabred}{\textbf{61.99}} & \textcolor{tabred}{\textbf{76.53}} & 55.23 & \textcolor{tabblue}{\textbf{64.42}} & \textcolor{tabred}{\textbf{42.32}} & \textcolor{tabred}{\textbf{59.47}} & \textcolor{tabred}{\textbf{64.38}} & \textcolor{tabblue}{\textbf{63.25}} & \textcolor{tabred}{\textbf{46.85}} & \textcolor{tabred}{\textbf{63.81}} \\
    \midrule
    \cellcolor{gray!10}Fully-Supervised~\cite{fang2021snunet} & \cellcolor{gray!10}92.80 & \cellcolor{gray!10}89.60 & \cellcolor{gray!10}83.80 & \cellcolor{gray!10}91.20 & \cellcolor{gray!10}86.50 & \cellcolor{gray!10}81.90 & \cellcolor{gray!10}72.80 & \cellcolor{gray!10}84.12 & \cellcolor{gray!10}68.40 & \cellcolor{gray!10}70.30 & \cellcolor{gray!10}53.80 & \cellcolor{gray!10}69.26 & \cellcolor{gray!10}79.90 & \cellcolor{gray!10}66.20 & \cellcolor{gray!10}56.70 & \cellcolor{gray!10}72.40 \\
    \bottomrule
    \end{tabular}}
    \label{table6}
\end{table*}

\begin{figure*}[t]
  \centering
  \includegraphics[width=\linewidth]{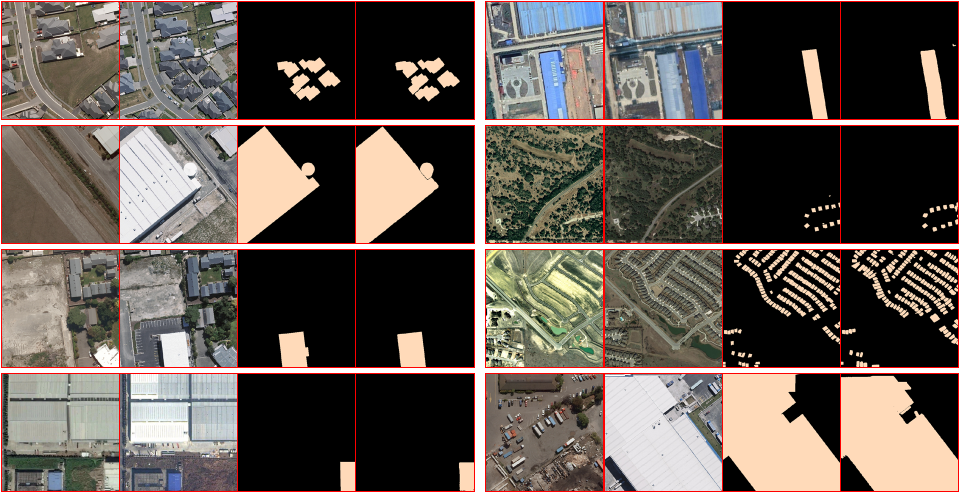}
  \caption{Qualitative results of AdaptOVCD on building change detection datasets including LEVIR-CD, WHU-CD, DSIFN, and the building category of SECOND. Each row displays a representative sample with varying building scales and densities. Columns from left to right: pre-change image $I_a$, post-change image $I_b$, ground truth annotation, and AdaptOVCD prediction. Color legend: \textcolor[RGB]{255,218,185}{\rule{0.8em}{0.8em}} Building change.}
  \label{fig3}
\end{figure*}

\begin{table*}[t]
\centering
\caption{Open-vocabulary change detection results on the SECOND dataset across six semantic categories. This evaluation assesses the framework's generalization to arbitrary change types specified via text prompts. \colorbox{blue!8}{IoU$^C$} is a standard metric. \colorbox{orange!10}{F$_1^C$} serves as the primary evaluation metric, with \textcolor{tabred}{\textbf{best}} results highlighted.}
\label{tab:second_classes}
\renewcommand{\arraystretch}{1.1}
\resizebox{\textwidth}{!}{
\begin{tabular}{l|cc|cc|cc|cc|cc|cc}
\toprule
& \multicolumn{12}{c}{\textbf{SECOND}} \\
\cmidrule(lr){2-13}
& \multicolumn{2}{c|}{\textbf{Building}} & \multicolumn{2}{c|}{\textbf{Low Vegetation}} & \multicolumn{2}{c|}{\textbf{Non-veg. Ground}} & \multicolumn{2}{c|}{\textbf{Playground}} & \multicolumn{2}{c|}{\textbf{Tree}} & \multicolumn{2}{c}{\textbf{Water}} \\
\textbf{Method} & \cellcolor{blue!8}IoU$^C$ & \cellcolor{orange!10}F$_1^C$ & \cellcolor{blue!8}IoU$^C$ & \cellcolor{orange!10}F$_1^C$ & \cellcolor{blue!8}IoU$^C$ & \cellcolor{orange!10}F$_1^C$ & \cellcolor{blue!8}IoU$^C$ & \cellcolor{orange!10}F$_1^C$ & \cellcolor{blue!8}IoU$^C$ & \cellcolor{orange!10}F$_1^C$ & \cellcolor{blue!8}IoU$^C$ & \cellcolor{orange!10}F$_1^C$ \\
\midrule
DynamicEarth (IMC)~\cite{li2025dynamicearth} & 32.03 & 48.51 & 0.40 & 0.79 & 0.00 & 0.00 & 26.93 & 42.43 & 10.64 & 19.23 & 12.34 & 21.97 \\
DynamicEarth (MCI)~\cite{li2025dynamicearth} & 37.76 & 54.28 & 21.21 & 34.99 & 24.18 & 38.95 & 17.44 & 29.70 & 12.48 & 22.26 & 13.49 & 23.77 \\
\midrule
AdaptOVCD (Ours) & \textcolor{tabred}{\textbf{46.85}} & \textcolor{tabred}{\textbf{63.81}} & \textcolor{tabred}{\textbf{22.15}} & \textcolor{tabred}{\textbf{36.27}} & \textcolor{tabred}{\textbf{33.99}} & \textcolor{tabred}{\textbf{50.74}} & \textcolor{tabred}{\textbf{29.28}} & \textcolor{tabred}{\textbf{45.30}} & \textcolor{tabred}{\textbf{12.67}} & \textcolor{tabred}{\textbf{22.49}} & \textcolor{tabred}{\textbf{23.34}} & \textcolor{tabred}{\textbf{37.84}} \\
\bottomrule
\end{tabular}}
\end{table*}

\textbf{Evaluation Metrics.} We employ standard evaluation metrics in the change detection field. All metrics are computed for the Changed class, denoted by the superscript $^C$. Precision$^C$ measures the proportion of true changes among pixels predicted as changed. Recall$^C$ measures the proportion of correctly detected pixels among all true change pixels. F$_1^C$ is the harmonic mean of the two, comprehensively reflecting detection accuracy and completeness. IoU$^C$ calculates the intersection-over-union of predicted and ground truth change regions, imposing more stringent requirements on boundary precision. Since F$_1^C$ balances precision and recall, we use it as the primary evaluation metric.

\subsection{Comparison with State-of-the-Art Methods}
To quantitatively assess the detection performance of AdaptOVCD, we first conduct comparative experiments in the well-established domain of building change detection. Since AdaptOVCD is a training-free open-vocabulary method, to ensure fair comparison, we select unsupervised methods and training-free methods for comparison. Unsupervised methods include CVA~\cite{4039609}, IRMAD~\cite{4060945}, PCA-Kmeans~\cite{5196726}, ISFA~\cite{isfa}, DSFA~\cite{8824216}, and DCVA~\cite{8608001}, which do not require labeled data but require training on the target data. Training-free methods include AnyChange~\cite{NEURIPS2024_94154162}, which generates binary change masks by mining feature similarities in the SAM latent space. UCD-SCM~\cite{10642429} combines the multi-scale features of FastSAM with the semantic filtering capability of CLIP to identify changes. DynamicEarth~\cite{li2025dynamicearth} systematically summarizes the OVCD task, and we selected the MCI architecture comprising SAM, DINOv2, and SegEarth-OV, and the IMC architecture consisting of APE and DINOv2, for comparison. Comparisons are conducted on three building change detection datasets (LEVIR-CD, WHU-CD, DSIFN) and the building category of the SECOND dataset, totaling four change scenarios. We also report the best performance of fully-supervised methods on each dataset as an upper-bound reference, which is not included in the method comparison.

Table~\ref{table6} presents the quantitative comparison results. Traditional unsupervised methods like CVA and IRMAD show high recall on some metrics but generally are restricted by low precision due to the lack of semantic understanding, making it difficult to distinguish semantic changes from extraneous variations such as lighting or seasonal changes. For instance, although ISFA achieves high recall, its extremely unbalanced precision indicates that it misclassifies a large number of background regions as changes. Existing training-free methods like AnyChange and UCD-SCM introduce generalized features from foundation models, significantly improving performance over traditional methods. However, limited by simple feature interaction modes, their overall F$_1^C$ still remains constrained by bottlenecks. DynamicEarth's MCI and IMC architectures achieve some performance gains but still fall short of the theoretical potential achievable by training-free methods, as their multi-stage model cascading leads to performance bottlenecks.

In contrast, AdaptOVCD exhibits superior detection robustness through its multi-level adaptive information fusion mechanism, especially in identifying man-made objects like buildings, substantially surpassing and being more stable than DynamicEarth's two architectures. Quantitative results show that AdaptOVCD achieves the best performance among existing unsupervised and training-free methods across all four datasets. Taking F$_1^C$ as an example, AdaptOVCD achieves 68.00\%, 76.53\%, 59.47\%, and 63.81\% on LEVIR-CD, WHU-CD, DSIFN, and SECOND (Building), respectively, improving by 2.22\%, 6.85\%, 4.49\%, and 9.53\% compared to the second-best training-free method. Notably, without any labeled samples, AdaptOVCD averages 84.89\% of the fully-supervised performance upper bound across the four building detection scenarios, demonstrating its significant potential for application in label-scarce scenarios.

Considering the generally suboptimal visual results of existing unsupervised and training-free methods in most scenarios, Figure~\ref{fig3} only presents qualitative results of AdaptOVCD on building change detection tasks to underscore its superiority. We selected representative building change samples with varying scales from the four datasets mentioned above, covering dense residential areas to large industrial plants. It can be seen that in this area, where building change features are relatively significant, AdaptOVCD demonstrates excellent performance, accurately capturing building change targets of different scales and generating detection masks with clear boundaries that are highly consistent with ground truth labels.

\begin{figure*}[t]
  \centering
  \includegraphics[width=\linewidth]{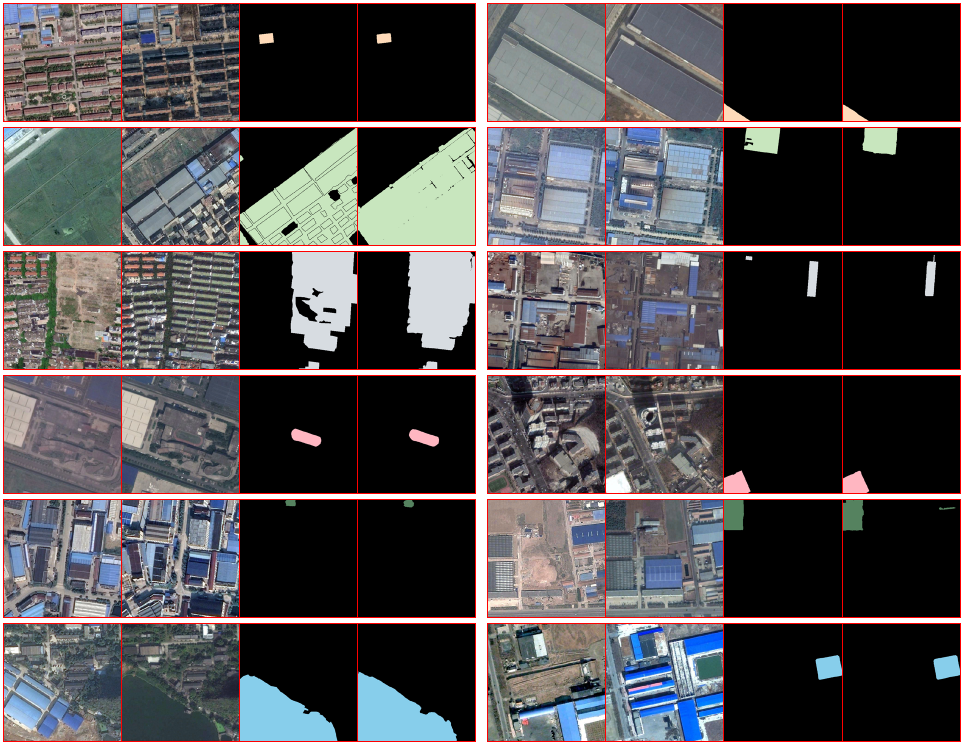}
  \caption{Open-vocabulary change detection results on the SECOND dataset across six semantic categories. Each row demonstrates detection capability for a specific land-cover change type driven by text prompts. Columns from left to right: pre-change image $I_a$, post-change image $I_b$, ground truth annotation, and AdaptOVCD prediction. Color legend: \textcolor[RGB]{255,218,185}{\rule{0.8em}{0.8em}} Building, \textcolor[RGB]{200,230,190}{\rule{0.8em}{0.8em}} Low Vegetation, \textcolor[RGB]{215,220,225}{\rule{0.8em}{0.8em}} Non-vegetated Ground, \textcolor[RGB]{255,182,193}{\rule{0.8em}{0.8em}} Playground, \textcolor[RGB]{85,130,95}{\rule{0.8em}{0.8em}} Tree, \textcolor[RGB]{135,206,235}{\rule{0.8em}{0.8em}} Water.}
  \label{fig4}
\end{figure*}

\subsection{OVCD Evaluation}

To evaluate AdaptOVCD's capability in open-vocabulary scenarios, we utilize the multi-class characteristic of the SECOND dataset to simulate an open-world detection task. This task mandates the model to identify six types of land cover changes, including buildings, low vegetation, non-vegetated ground surfaces, playgrounds, trees, and water, based solely on text instructions. Here, we benchmark AdaptOVCD against DynamicEarth, the only currently available open-source OVCD baseline method. In the experiment, each method employs the predefined text descriptions in Table~\ref{tab:prompts} as input and performs zero-shot inference without any access to validation set ground truth, validating its ability to comprehend and potentialize different semantic changes.

Table~\ref{tab:second_classes} presents the quantitative evaluation results, demonstrating that AdaptOVCD significantly outperforms the existing training-free method DynamicEarth in both detection accuracy and category adaptability. Notably, DynamicEarth's IMC architecture achieves near-zero performance on certain categories, consistent with the original paper's observations, stemming from its difficulty in instantiating abstract background category descriptions. The MCI architecture circumvents this issue by adopting a segment-then-identify paradigm similar to ours, yet AdaptOVCD exhibits more stable detection performance attributable to its dual-dimensional multi-level adaptive information fusion mechanism. The data reveal that AdaptOVCD not only improves accuracy by 9.53\% over the MCI architecture on the Building category but also maintains more balanced and robust performance across various complex land cover scenarios, validating the generalization advantage of the proposed framework in open-world settings.

The detection performance for each category, quantified by F$_1^C$, exhibits significant variance, reflecting the inherent adaptability laws when vision foundation models work synergistically. For Buildings, the F$_1^C$ reaches as high as 63.81\%, with this optimal performance driven by regular geometric structures and clear physical boundaries providing ideal segmentation conditions for SAM-HQ, and significant man-made object features facilitating precise recognition by DGTRS-CLIP. Secondly, Non-vegetated Ground Surface and Playground achieve F$_1^C$ of 50.74\% and 45.30\% respectively, showing good detection robustness due to their significant separability from natural backgrounds in texture. It is noteworthy that although Water and Low Vegetation achieve F$_1^C$ of 37.84\% and 36.27\% respectively, their detection accuracy is at a medium level due to dynamic water level changes and spectral confusion with surrounding vegetation. In contrast, Tree achieves an F$_1^C$ of only 22.49\%, constituting a detection challenge. This performance degradation stems partly from semantic ambiguity caused by natural overlap with low vegetation in spectral space. More fundamentally, the non-rigid boundaries and shadow interference of tree crowns in overhead views cause severe boundary mismatches during the instance segmentation stage, thereby impeding correct semantic association.

Figure~\ref{fig4} further displays the visualization results of AdaptOVCD on the six categories mentioned above. As observed in the figure, AdaptOVCD keenly captures change features of different semantic categories. Specifically, the model accurately identifies new construction and demolition of buildings, effectively distinguishes subtle changes between grass and shrubs, and precisely locates the expansion/shrinkage of water bodies and changes in sports fields under complex backgrounds through text guidance. These qualitative results are highly consistent with quantitative metrics, intuitively proving the generalization ability of the AdaptOVCD framework to achieve arbitrary category change detection driven by text.

\subsection{Ablation Study}

To comprehensively evaluate the effectiveness of the dual-dimensional multi-level information fusion mechanism, we conducted extensive ablation experiments. Starting with the complete AdaptOVCD framework, we examined their contributions by sequentially removing ARA, ACT, and ACF. When ARA is removed, original image pairs are used directly. When ACT is removed, a fixed percentile threshold is used for change determination. When ACF is removed, the raw classification results of DGTRS-CLIP are output directly.

Table~\ref{table:ablation} presents the quantitative results, using the average F$_1^C$ across nine scenarios as the metric. The complete framework reaches 51.16\%. When a single module is removed, removing ACF drops performance by 1.99\%, removing ACT drops by 1.35\%, and removing ARA drops by 1.00\%, indicating that the contributions of ACF and ACT are significantly higher than ARA. Performance drops further when two modules are removed. Specifically, retaining only ARA drops performance to 47.45\%, a loss of 3.71\%. Retaining only ACT drops it to 48.30\%, a drop of 2.86\%. Retaining only ACF drops it to 48.48\%, a drop of 2.68\%. Removing all modules drops performance to 47.00\%, a 4.16\% decline compared to the complete framework.

Analyzing these trends from an information fusion perspective, ARA, as a data-level design unifying the radiometric characteristics of bi-temporal images, lays a signal consistency foundation for subsequent processing. Its contribution is relatively limited but indispensable. ACT and ACF act directly on the core reasoning process. Specifically, ACT dynamically reconstructs decision boundaries by fusing global difference distributions with edge structure priors, while ACF calibrates decision outputs by combining semantic confidence with spatial consistency constraints. Both effectively promote deep synergy among heterogeneous models, thus contributing more significantly to overall performance. The experimental results fully confirm the necessity of the dual-dimensional multi-level information fusion strategy.

\begin{table*}[t]
    \centering
    \caption{Ablation study evaluating the contribution of each adaptive module across nine change detection scenarios. The baseline removes all three modules: ARA is disabled with direct image input, ACT is replaced by a fixed percentile threshold for change determination, and ACF is bypassed to directly output DGTRS-CLIP predictions without confidence filtering. F$_1^C$ (\%) is reported, with \textcolor{tabred}{\textbf{best}} results highlighted.}
    \renewcommand{\arraystretch}{1.1}
    \resizebox{\textwidth}{!}{
    \begin{tabular}{ccc|ccc|cccccc}
        \toprule
        \multicolumn{3}{c|}{\textbf{Modules}} & \multicolumn{3}{c|}{\textbf{Building}} & \multicolumn{6}{c}{\textbf{SECOND}} \\
        \cmidrule(lr){1-3} \cmidrule(lr){4-6} \cmidrule(lr){7-12}
        \textbf{ARA} & \textbf{ACT} & \textbf{ACF} & \textbf{LEVIR-CD} & \textbf{WHU-CD} & \textbf{DSIFN} & \textbf{Building} & \textbf{Low Veg.} & \textbf{Non-veg.} & \textbf{Playground} & \textbf{Tree} & \textbf{Water} \\
        \midrule
        $\times$ & $\times$ & $\times$ & 65.97 & 71.36 & 55.82 & 59.68 & 33.55 & 46.18 & 40.18 & 19.15 & 31.15 \\
        \checkmark & $\times$ & $\times$ & 66.29 & 72.13 & 56.54 & 60.09 & 33.90 & 46.85 & 39.75 & 19.71 & 31.76 \\
        $\times$ & \checkmark & $\times$ & 66.90 & 72.91 & 56.96 & 61.34 & 34.03 & 47.13 & 42.62 & 20.33 & 32.51 \\
        $\times$ & $\times$ & \checkmark & 66.75 & 73.43 & 57.20 & 61.79 & 34.31 & 47.50 & 41.48 & 21.06 & 32.77 \\
        \checkmark & \checkmark & $\times$ & 66.97 & 74.90 & 57.78 & 62.43 & 34.72 & 48.25 & 42.78 & 21.51 & 33.19 \\
        \checkmark & $\times$ & \checkmark & 67.19 & 75.61 & 57.94 & 62.90 & 35.31 & 48.87 & 43.35 & 21.30 & 35.86 \\
        $\times$ & \checkmark & \checkmark & 67.56 & 75.73 & 58.23 & 62.93 & 35.67 & 49.03 & 43.62 & 21.93 & 36.73 \\
        \checkmark & \checkmark & \checkmark & \textcolor{tabred}{\textbf{68.00}} & \textcolor{tabred}{\textbf{76.53}} & \textcolor{tabred}{\textbf{59.48}} & \textcolor{tabred}{\textbf{63.81}} & \textcolor{tabred}{\textbf{36.27}} & \textcolor{tabred}{\textbf{50.74}} & \textcolor{tabred}{\textbf{45.30}} & \textcolor{tabred}{\textbf{22.49}} & \textcolor{tabred}{\textbf{37.84}} \\
        \bottomrule
    \end{tabular}}
    \label{table:ablation}
\end{table*}

\begin{figure*}[t]
  \centering
  \includegraphics[width=\linewidth]{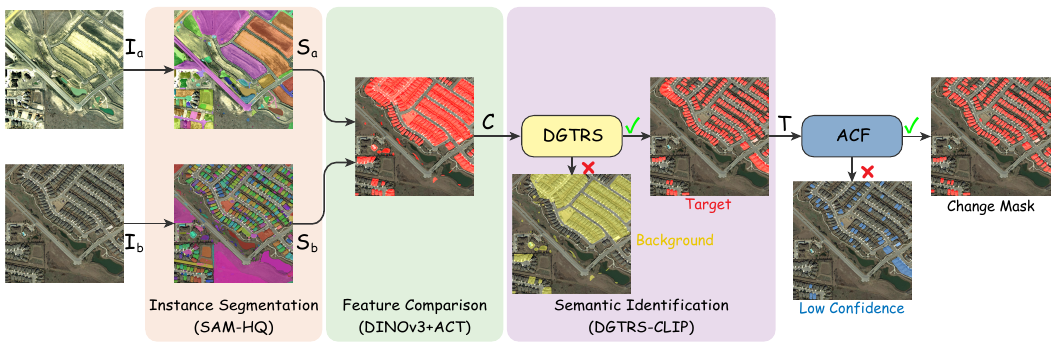}
  \caption{Intermediate process visualization illustrating the progressive filtering mechanism of AdaptOVCD on two representative LEVIR-CD samples. The top row shows a true positive case where building changes are successfully detected through the three-stage pipeline. The bottom row demonstrates effective false positive suppression, where spurious detections are progressively eliminated. Processing stages are color-coded: light orange for Instance Segmentation ($S_a$, $S_b$), light green for Feature Comparison (candidates $C$), and light purple for semantic identification with yellow DGTRS-CLIP and blue ACF. Symbols: \textcolor{green}{\checkmark} accepted, \textcolor{red}{$\times$} rejected by DGTRS-CLIP, blue regions filtered by ACF due to low confidence.}
  \label{fig5}
\end{figure*}

\subsection{Intermediate Process Visualization}

To gain a deeper understanding of the mechanism of each module, Figure~\ref{fig5} presents visualizations of the intermediate processes of AdaptOVCD processing two typical pairs of samples on the LEVIR-CD dataset. ARA, as a data-level preprocessing step, is not shown in the figure. In the light orange region, SAM-HQ performs dense instance segmentation on the bi-temporal images $I_a$ and $I_b$, generating masks $S_a$ and $S_b$, with different colors in the figure identifying each instance region. In the light green region, DINOv3 extracts bi-temporal semantic features, and ACT outputs change candidates $C$ through adaptive thresholds, with red masks identifying all detected change regions. In the light purple region, DGTRS-CLIP semantically classifies candidate masks based on mutually exclusive Target and Background text prototypes, retaining red masks identified as Target and discarding yellow masks identified as Background. Finally, ACF fuses confidence and spatial consistency constraints to perform instance-level filtering, discarding low-confidence blue masks and outputting high-confidence red masks as the detection result.

Comparing the two sets of samples, one can observe AdaptOVCD's progressive filtering effect. The top row sample demonstrates the correct detection of a newly constructed building scene. SAM-HQ accurately segments the boundaries of dense residential buildings. After identification by DINOv3+ACT, semantic confirmation by DGTRS-CLIP, and confidence filtering by ACF, precise change masks are ultimately output. The bottom row sample demonstrates the effective suppression of false positives. Although DINOv3+ACT detects several change candidates, DGTRS-CLIP classifies most as Background and discards them, and ACF further filters low-confidence regions, ultimately avoiding false alarms. The visualization process above intuitively verifies the effectiveness of the dual-dimensional multi-level information fusion architecture, ensuring the reliability of detection results.

\begin{table*}[t]
\centering
\caption{Architecture ablation study comparing different foundation model combinations by replacing SAM-HQ and DINOv3 with their earlier versions. The semantic identification stage uniformly uses DGTRS-CLIP. F$_1^C$ (\%) is reported, with \textcolor{tabred}{\textbf{best}} results highlighted.}
\label{tab:architecture}
\renewcommand{\arraystretch}{1.15}
\resizebox{\textwidth}{!}{
\begin{tabular}{cc|cccc|cccccc}
\toprule
\multicolumn{2}{c|}{\textbf{Architecture}} & \multicolumn{3}{c|}{\textbf{Building}} & \multicolumn{6}{c}{\textbf{SECOND}} \\
\cmidrule(lr){1-2} \cmidrule(lr){3-5} \cmidrule(lr){6-11}
\textbf{Segmentor} & \textbf{Comparator} & \textbf{LEVIR-CD} & \textbf{WHU-CD} & \textbf{DSIFN} & \textbf{Building} & \textbf{Low Veg.} & \textbf{Non-veg.} & \textbf{Playground} & \textbf{Tree} & \textbf{Water} \\
\midrule
SAM & DINO & 63.19 & 65.80 & 50.85 & 58.15 & 31.53 & 45.11 & 35.69 & 16.64 & 27.52 \\
SAM & DINOv2 & 66.07 & 76.21 & 57.98 & 62.68 & 35.41 & 50.55 & 44.65 & 17.53 & 31.14 \\
SAM & DINOv3 & 66.73 & 76.93 & 58.29 & 64.97 & 36.01 & 50.49 & 42.16 & 21.14 & 32.15 \\
\midrule
SAM-HQ & DINO & 64.70 & 66.91 & 49.95 & 57.53 & 31.38 & 44.58 & 33.74 & 18.09 & 33.16 \\
SAM-HQ & DINOv2 & 67.37 & 75.92 & 58.20 & 60.96 & 34.02 & 49.57 & 42.03 & 18.06 & 36.43 \\
\midrule
SAM-HQ & DINOv3 & \textcolor{tabred}{\textbf{68.00}} & \textcolor{tabred}{\textbf{76.53}} & \textcolor{tabred}{\textbf{59.47}} & \textcolor{tabred}{\textbf{63.81}} & \textcolor{tabred}{\textbf{36.27}} & \textcolor{tabred}{\textbf{50.74}} & \textcolor{tabred}{\textbf{45.30}} & \textcolor{tabred}{\textbf{22.49}} & \textcolor{tabred}{\textbf{37.84}} \\
\bottomrule
\end{tabular}}
\end{table*}

\subsection{Architecture Ablation}

To explore the impact of foundation model performance on the AdaptOVCD multi-level architecture, we conducted replacement experiments on models in the instance segmentation and feature comparison stages under conditions of similar model scale. Results are detailed in Table~\ref{tab:architecture}. The instance segmentation stage compares SAM-HQ with SAM, and the feature comparison stage compares DINOv3, DINOv2, and DINO. It is worth noting that SAM and SAM-HQ, as well as the DINO-series models used in the experiments, all have pre-trained weights of equivalent parameter scale. This replacement strategy ensures minimal interference factors. DINO series improvements are mainly reflected in stronger semantic extraction capabilities, while SAM-HQ's advantage over SAM lies mainly in more precise segmentation of blurred boundaries. Since the text prototypes in Table~\ref{tab:prompts} are optimized for the semantic space of DGTRS-CLIP, the semantic identification stage is not included in the replacement scope.

Analyzing from the instance segmentation stage, the performance improvement of SAM-HQ compared to SAM does not hold in all scenarios. Its core advantage lies in the precise segmentation of blurred boundaries. For the Water category, SAM-HQ with DINOv3 achieves 37.84\%, significantly better than SAM with DINOv3's 32.15\%, benefiting from the HQ-Token mechanism's precise capture of blurred boundaries like water edges. Although in the Building category, fine segmentation might affect spatial correspondence due to excessive fragmentation, the combination of SAM-HQ and DINOv3 still achieves optimal performance in most categories, indicating that high-quality boundary segmentation is an important foundation for improving overall performance.

Analyzing from the feature comparison stage, DINO-series models show clear generational performance improvements, indicating that the stronger the feature extraction capability, the stronger the model's overall detection performance. Taking SAM as an example, DINO, DINOv2, and DINOv3 show continuous improvement in detection performance on the Building task. It should be noted that the DINO model itself does not possess change judgment capability but assists the subsequent ACT module in constructing more robust change decision boundaries by providing more semantically discriminative feature representations. Therefore, DINOv3's stronger feature representation capability directly translates into more accurate change recognition in downstream tasks.

\subsection{Cross-Dataset Evaluation}

To verify AdaptOVCD's generalization ability, we selected the most representative building change detection scenarios. Table~\ref{table:cross_dataset} compares the training-free AdaptOVCD with fully-supervised models trained on different datasets in terms of cross-dataset performance. Gray diagonal cells in the table indicate the best performance of fully-supervised models on the same dataset, while off-diagonal cells show cross-dataset test results.

From the table, it can be seen that fully-supervised models exhibit severe performance degradation during cross-dataset evaluation. For example, the LEVIR-CD trained model achieves an F$_1^C$ of 91.20\% when tested on the same dataset, but drops to 51.23\%, 21.23\%, and 27.80\% on WHU-CD, DSIFN, and SECOND, respectively. Some scenarios even present extreme Precision and Recall imbalances. For instance, the DSIFN trained model achieves a Recall of 64.90\% on SECOND but a Precision of only 33.30\%. This imbalance stems from the difficulty of decision boundaries learned in the source domain to adapt to the change feature distribution in the target domain.

In contrast, AdaptOVCD demonstrates significant cross-dataset stability. Across the four datasets, AdaptOVCD's F$_1^C$ values are 68.00\%, 76.53\%, 59.47\%, and 63.81\%, respectively, with minimal performance fluctuations, and it maintains a good balance between Precision and Recall on all datasets, avoiding the extreme imbalance problem of fully-supervised models across domains, thereby displaying strong versatility. This validates that the advantage of AdaptOVCD's adoption of a training-free architecture lies in its ability to achieve flexible detection for different change detection scenarios simply by employing adapted detection vocabulary.

\begin{table*}[t]
    \centering
    \caption{Cross-dataset generalization evaluation for building change detection. Each row indicates the training source, and columns show test performance on different target datasets. Precision (\colorbox{blue!8}{Prec.}), Recall (\colorbox{blue!8}{Rec.}) and \colorbox{blue!8}{IoU$^C$} are standard metrics. \colorbox{orange!10}{F$_1^C$} serves as the primary evaluation metric. Fully-supervised models exhibit severe performance degradation when tested on unseen datasets, while AdaptOVCD maintains consistent performance without any training. \colorbox{gray!10}{Gray cells} denote in-domain supervised upper bounds (excluded from ranking). \textcolor{tabred}{\textbf{Best}} and \textcolor{tabblue}{\textbf{second best}} cross-dataset results are highlighted.}
    \renewcommand{\arraystretch}{1.1}
    \resizebox{\textwidth}{!}{
    \begin{tabular}{l|cccc|cccc|cccc|cccc}
    \toprule
    \diagbox{Train on:}{Test on:} & \multicolumn{4}{c|}{\textbf{LEVIR-CD}} & \multicolumn{4}{c|}{\textbf{WHU-CD}} & \multicolumn{4}{c|}{\textbf{DSIFN}} & \multicolumn{4}{c}{\textbf{Building on SECOND}} \\
    \cmidrule(lr){2-5} \cmidrule(lr){6-9} \cmidrule(lr){10-13} \cmidrule(lr){14-17}
    & \cellcolor{blue!8}Prec. & \cellcolor{blue!8}Rec. & \cellcolor{blue!8}IoU$^C$ & \cellcolor{orange!10}F$_1^C$ & \cellcolor{blue!8}Prec. & \cellcolor{blue!8}Rec. & \cellcolor{blue!8}IoU$^C$ & \cellcolor{orange!10}F$_1^C$ & \cellcolor{blue!8}Prec. & \cellcolor{blue!8}Rec. & \cellcolor{blue!8}IoU$^C$ & \cellcolor{orange!10}F$_1^C$ & \cellcolor{blue!8}Prec. & \cellcolor{blue!8}Rec. & \cellcolor{blue!8}IoU$^C$ & \cellcolor{orange!10}F$_1^C$ \\
    \midrule
    LEVIR-CD & \cellcolor{gray!10}92.80 & \cellcolor{gray!10}89.60 & \cellcolor{gray!10}83.80 & \cellcolor{gray!10}91.20 & \textcolor{tabblue}{\textbf{75.30}} & \textcolor{tabblue}{\textbf{42.10}} & \textcolor{tabblue}{\textbf{36.50}} & \textcolor{tabblue}{\textbf{51.23}} & 52.50 & 19.50 & 13.80 & 21.23 & \textcolor{tabblue}{\textbf{61.00}} & 18.00 & 16.10 & 27.80 \\
    WHU-CD & 48.20 & 31.80 & 24.60 & 37.45 & \cellcolor{gray!10}86.50 & \cellcolor{gray!10}81.90 & \cellcolor{gray!10}72.80 & \cellcolor{gray!10}84.12 & 15.20 & 23.00 & 10.05 & 18.26 & 16.50 & 25.80 & 10.87 & 19.87 \\
    DSIFN & 26.60 & \textcolor{tabblue}{\textbf{63.30}} & 23.00 & 37.50 & 14.80 & 23.50 & 10.02 & 18.21 & \cellcolor{gray!10}68.40 & \cellcolor{gray!10}70.30 & \cellcolor{gray!10}53.80 & \cellcolor{gray!10}69.26 & 33.30 & \textcolor{tabred}{\textbf{64.90}} & \textcolor{tabblue}{\textbf{28.20}} & \textcolor{tabblue}{\textbf{44.00}} \\
    SECOND & \textcolor{tabblue}{\textbf{60.70}} & 53.20 & \textcolor{tabblue}{\textbf{39.60}} & \textcolor{tabblue}{\textbf{56.70}} & 18.00 & 27.50 & 12.28 & 21.89 & \textcolor{tabred}{\textbf{83.00}} & \textcolor{tabblue}{\textbf{31.50}} & \textcolor{tabblue}{\textbf{29.60}} & \textcolor{tabblue}{\textbf{45.60}} & \cellcolor{gray!10}79.90 & \cellcolor{gray!10}66.20 & \cellcolor{gray!10}56.70 & \cellcolor{gray!10}72.40 \\
    \midrule
    AdaptOVCD (Ours) & \textcolor{tabred}{\textbf{62.83}} & \textcolor{tabred}{\textbf{74.10}} & \textcolor{tabred}{\textbf{51.52}} & \textcolor{tabred}{\textbf{68.00}} & \textcolor{tabred}{\textbf{80.60}} & \textcolor{tabred}{\textbf{72.85}} & \textcolor{tabred}{\textbf{61.99}} & \textcolor{tabred}{\textbf{76.53}} & \textcolor{tabblue}{\textbf{55.23}} & \textcolor{tabred}{\textbf{64.42}} & \textcolor{tabred}{\textbf{42.32}} & \textcolor{tabred}{\textbf{59.47}} & \textcolor{tabred}{\textbf{64.38}} & \textcolor{tabblue}{\textbf{63.25}} & \textcolor{tabred}{\textbf{46.85}} & \textcolor{tabred}{\textbf{63.81}} \\
    \bottomrule
    \end{tabular}}
    \label{table:cross_dataset}
\end{table*}

\section{Conclusion}
\label{sec:conclusion}

In this paper, we propose AdaptOVCD, a training-free open-vocabulary change detection framework for remote sensing imagery. By decomposing the complex OVCD task into three sub-tasks and introducing domain-specific foundation models, this method effectively bridges the semantic gap of general vision models in overhead perspectives. Meanwhile, a dual-dimensional multi-level information fusion architecture is constructed, achieving cascaded information fusion at the data, feature, and decision levels vertically, and systematically suppressing error accumulation through targeted adaptive designs horizontally. Extensive experiments covering nine scenarios demonstrate that AdaptOVCD not only achieves an average of 84.89\% of the fully-supervised performance upper bound in building detection but also possesses stronger generalization capabilities, confirming its immense potential in open-world applications.

Future work will focus on two main aspects. On the one hand, we will commit to further tapping into the potential of the training-free architecture, exploring lightweight or more powerful foundation model combinations. On the other hand, the zero-shot detection capability of AdaptOVCD provides an efficient pre-annotation means for constructing large-scale OVCD labeled datasets, facilitating the rapid accumulation of high-quality samples combined with manual correction, thereby laying a data foundation for the training of end-to-end fully supervised OVCD models.

\section*{Data Availability}
The datasets used in this study are publicly available: LEVIR-CD, WHU-CD, DSIFN, and SECOND.

\section*{Authorship Contribution Statement}
Mingyu Dou: Writing – original draft, Methodology, Software, Visualization; Shi Qiu: Supervision, Methodology, Writing – review \& editing; Ming Hu: Validation, Writing – review \& editing; Yifan Chen: Writing – review \& editing; Huping Ye: Writing – review \& editing; Xiaohan Liao: Writing – review \& editing; Zhe Sun: Supervision, Conceptualization, Writing – review \& editing.

\section*{Declaration of Competing Interest}
The authors declare that they have no known competing financial interests or personal relationships that could have appeared to influence the work reported in this paper. 

\section*{Acknowledgments}
This work was supported by the National Key R\&D Program of China (2023YFB3905700) and the Shenzhen Science and Technology Program (KJZD20230923115210021).

\printcredits

%\bibliographystyle{unsrt}
\bibliographystyle{elsarticle-num}
%\bibliographystyle{cas-model2-names}

\bibliography{ref}

\end{document}